\def\year{2021}\relax
\documentclass[letterpaper]{article} % DO NOT CHANGE THIS
\usepackage{aaai21}  % DO NOT CHANGE THIS
\usepackage{times}  % DO NOT CHANGE THIS
\usepackage{helvet} % DO NOT CHANGE THIS
\usepackage{courier}  % DO NOT CHANGE THIS
\usepackage[hyphens]{url}  % DO NOT CHANGE THIS
\usepackage{graphicx} % DO NOT CHANGE THIS
\usepackage{natbib}  % DO NOT CHANGE THIS AND DO NOT ADD ANY OPTIONS TO IT
\usepackage{caption} % DO NOT CHANGE THIS AND DO NOT ADD ANY OPTIONS TO IT
\usepackage{adjustbox}
\usepackage{subfigure}
\usepackage{booktabs}

\newtheorem{definition}{Definition}

\title{On Baselines for Local Feature Attributions}
\author {
        Johannes Haug,
        Stefan Zürn,
        Peter El-Jiz,
        Gjergji Kasneci \\
}
\affiliations {
    University of Tuebingen, Germany \\
    \{johannes-christian.haug, gjergji.kasneci\}@uni-tuebingen.de;\{stefan.zuern, peter.el-jiz\}@student.uni-tuebingen.de
}

\begin{document}
%\linenumbers

\maketitle

\begin{abstract}
High-performing predictive models, such as neural nets, usually operate as black boxes, which raises serious concerns about their interpretability. Local feature attribution methods help to explain black box models and are therefore a powerful tool for assessing the reliability and fairness of predictions. To this end, most attribution models compare the importance of input features with a reference value, often called baseline. Recent studies show that the baseline can heavily impact the quality of feature attributions. Yet, we frequently find simplistic baselines, such as the zero vector, in practice. In this paper, we show empirically that baselines can significantly alter the discriminative power of feature attributions. We conduct our analysis on tabular data sets, thus complementing recent works on image data. Besides, we propose a new taxonomy of baseline methods. Our experimental study illustrates the sensitivity of popular attribution models to the baseline, thus laying the foundation for a more in-depth discussion on sensible baseline methods for tabular data.
\end{abstract}

\section{Introduction}
Neural nets and other complex predictive models perform well in a variety of applications. In practice, however, complex black-box models can give rise to serious concerns about interpretability. Feature weighting and local attribution methods help to explain complex models and thus improve the interpretability of predictions \cite{haug2020leveraging,kasneci2016licon,lundberg2017unified,ribeiro2016should,shrikumar2017learning,sundararajan2017axiomatic}. Accordingly, feature attributions are an important step towards more transparent and fair machine learning.

Local attribution methods usually assess the importance of features with respect to a reference input or baseline value \cite{sundararajan2019many,izzo2020baseline}. The baseline is closely related to the concept of missingness, i.e. the (approximate) neutral value that a feature would take, if it were considered missing. For example, \citet{lundberg2017unified} replace the missing values in a sampled feature coalition with their expected value and \citet{sundararajan2017axiomatic} propose to use a black image as baseline in object recognition tasks.

The concept of missingness is domain-specific and may therefore vary in practice. For example, suppose $x_i \in [0,255]$ denotes one pixel of an image. Here, $x_i=0$ is a black pixel, which may indeed be considered missing information. However, suppose that $x_i \in \{0,1,2,3\}$ is a nominal feature. In this case, $x_i=0$ does not represent missingness. Accordingly, the zero baseline considered in this example would render the generated attributions meaningless. Indeed, several authors have recently expressed concerns about the appropriateness of popular baseline methods \cite{sundararajan2019many,sturmfels2020visualizing,izzo2020baseline}. In practice, a baseline should be selected with respect to the data distribution at hand.

In this paper, we propose a novel taxonomy of baseline methods, which may help compare and select baselines in research and practice. Along the taxonomy, we briefly introduce common baseline methods \cite{lundberg2017unified,sundararajan2019many,izzo2020baseline,sturmfels2020visualizing}. In a next step, we investigate the effect of different baselines on the feature attributions generated by four state-of-the-art attribution models. Our experiments focus on tabular data sets, which are often used as benchmarks in the explainability and fairness literature. We thereby complement and extend a recent study of \citet{sturmfels2020visualizing}, which illustrates the effects of different baselines for the classification of images. Our results suggest that the baseline can have a dramatic impact on the discriminative quality of the generated feature attributions. Strikingly, there was no universally best-performing baseline method. However, certain baselines rarely yielded discriminative feature attributions, suggesting that they may not be suitable for tabular data.

In summary, our contribution is two-fold: We categorize existing baseline methods into a new taxonomy. Besides, we provide an experimental evaluation of common baselines using several attribution models and tabular data sets. Hence, our work may serve as a reference for a deeper discussion of baseline methods in the context of tabular data and a more principled choice of baselines in practical applications.

%--------------------- Baselines ------------------------------------%
\begin{table*}[t]
\caption{\textbf{Baseline Taxonomy.} We categorize common baseline methods according to the Definitions \ref{def:static} and \ref{def:deterministic}.}
    \label{tab:baselines}
    \centering
    \begin{adjustbox}{max width=\textwidth}
        \begin{tabular}{lcc}
        \toprule
        \textbf{Baseline Name} & \textbf{Static/Dynamic} & \textbf{Deterministic/Stochastic}\\ 
        \cmidrule(lr){1-1} \cmidrule(lr){2-2} \cmidrule(lr){3-3}
        Constant (e.g. zero baseline) & static & deterministic \\
        Maximum Distance \cite{sturmfels2020visualizing} & dynamic & deterministic \\
        Blurred \cite{fong2017interpretable,sturmfels2020visualizing} & dynamic & stochastic\\
        Gaussian \cite{smilkov2017smoothgrad,sturmfels2020visualizing} & dynamic & stochastic\\
        Uniform \cite{sturmfels2020visualizing} & dynamic & stochastic\\
        Expectation \cite{lundberg2017unified} & static & stochastic\\
        Neutral \cite{izzo2020baseline} & static & deterministic\\
        \bottomrule
        \end{tabular}
    \end{adjustbox}
\end{table*}
%--------------------- Baselines ------------------------------------%
\section{A Taxonomy of Baseline Methods}
\label{sec:taxonomy}
Many different baseline methods have been introduced in recent years \cite{sundararajan2019many,izzo2020baseline,sturmfels2020visualizing}. Yet, to the best of our knowledge, there is no generally accepted scheme for categorising and comparing these methods. To this end, we propose a novel taxonomy.

Let $x_i,x_j \sim X$ be two arbitrary observations. Suppose we apply a baseline method $B$ to obtain a baseline corresponding to each observation, i.e. $B(x_i,x_j) = b_{x_i}, b_{x_j}$. We then define
\begin{definition}[Static or Dynamic Baseline]
\label{def:static}
    A baseline method $B$ is static, if $\forall i,j: b_{x_i} = b_{x_j}$. Otherwise, $B$ is dynamic, i.e. $\exists i,j: b_{x_i} \neq b_{x_j}$.
\end{definition}
Intuitively, a static baseline method provides the same baseline value for every observation, whereas a dynamic baseline method may provide different values. 

Next, suppose that we run the baseline method $B(x_i)$ two times for the observation $x_i$, which corresponds to $B^1(x_i) = b^1_{x_i}$ (first run) and $B^2(x_i) = b^2_{x_i}$ (second run). We then define
\begin{definition}[Deterministic or Stochastic Baseline]
\label{def:deterministic}
    A baseline method $B$ is deterministic, if the probability $Pr(b^1_{x_i} = b^2_{x_i}) = 1$. Otherwise, $B$ is stochastic, i.e. $Pr(b^1_{x_i} = b^2_{x_i}) < 1$.
\end{definition}
Intuitively, a deterministic baseline method always produces the same baseline with respect to an observation $x_i$ whenever it is called. On the other hand, a stochastic baseline method may be subject to variation.

\subsection{Categorizing Common Baseline Methods}
Next, we briefly introduce some popular and recent baseline methods. Table \ref{tab:baselines} shows the categorization of all methods according to the taxonomy defined above (Definition \ref{def:static}-\ref{def:deterministic}).

The \emph{constant} baseline is a prominent static and deterministic baseline. As the name suggests, the constant baseline is a fixed value that is specified once. Note that the zero baseline (i.e. black image) mentioned earlier is an instantiation of the constant baseline.

The \emph{maximum distance} baseline corresponds to an observation that is furthest away from the observation in question by the $\ell 1$-norm \cite{sturmfels2020visualizing}. Notably, the maximum distance baseline is dynamic and deterministic.

The \emph{blurred} baseline was originally introduced for image data \cite{sturmfels2020visualizing}. Specifically, this baseline method applies a Gaussian blur filter to the observation in question \cite{fong2017interpretable}. Note that the filter blurs each input feature with respect to its adjacent features. Accordingly, the blurred baseline requires an inherent sense of neighbourhood among input features, which might not always be evident in tabular data. In general, however, we may apply the blurred baseline to (numeric) tabular data as well. The blurred baseline method is dynamic and stochastic.

Similar to the blurred baseline, the \emph{Gaussian} baseline introduces noise to the original observation. To this end, one specifies a Gaussian distribution per input feature, which is centered at the original input value \cite{smilkov2017smoothgrad,sturmfels2020visualizing}. One then draws random samples from the generated distributions. Hence, the Gaussian baseline is dynamic and stochastic.

Likewise, we may draw random samples from uniform distributions per input feature. The uniform distributions are defined in the valid range of the original features \cite{sturmfels2020visualizing}. Again, the \emph{uniform} baseline is dynamic and stochastic.

\citet{lundberg2017unified} specified the baseline as a function of the expectation of a reference sample. We call this the \emph{expectation} baseline, which is static and stochastic as the reference sample is usually drawn randomly from the training data.

Finally, \citet{izzo2020baseline} argue that a baseline should lie on the decision boundary of the predictive model. Given that the decision boundary does not shift, this \emph{neutral} baseline is static and deterministic. At the time of writing this paper, the neutral baseline was only specified for certain neural network architectures and no open source implementation was available. For this reason, we did not consider the neutral baseline in our experiments.
% Nevertheless, the neutral baseline is worth mentioning, because it is a very intuitive approach that guarantees that the baseline is located in the original feature space of a data set.

%--------------------- Datasets ------------------------------------%
\begin{table*}[t]
\caption{\textbf{Data Sets.} ``Feature Types'' describes the data types included in the data set (cont. = continuous, cat. = categorical). ``Class Imbalance'' indicates whether the target distribution is imbalanced. For imbalanced data sets, we indicate the proportion of observations that belong to the positive class in parentheses. Note that in the Communities data, we have removed features with a high proportion of missing values to ensure valid evaluations. Besides, we have obtained a random sample of 50,000 observations from the Fraud Detection data set.}
    \label{tab:datasets}
    \centering
    \begin{adjustbox}{max width=\textwidth}
        \begin{tabular}{llllll}
        \toprule
        \textbf{Dataset} & \textbf{\# Observations} & \textbf{\# Features} & \textbf{Feature Types} & \textbf{\# Classes} & \textbf{Class Imbalance} \\ 
        \cmidrule(lr){1-1} \cmidrule(lr){2-6}
        Human Activity Recognition \cite{Dua:2019} & 10,299 & 561 & cont. & 6 & no \\
        Fraud Detection \cite{dal2014learned} & 50,000 & 30 & cont. & 2 & yes (99.8\% pos. class)\\
        Communities \cite{Dua:2019} & 1,993 & 100 & cat./cont. & 2 & yes (72\% pos. class)\\
        Spambase \cite{Dua:2019} & 4,601 & 57 & cont. & 2 & no\\
        COMPAS \cite{angwin2016machine} & 7,214 & 11 & cat./cont. & 2 & no\\
        \bottomrule
        \end{tabular}
    \end{adjustbox}
\end{table*}
%--------------------- Datasets ------------------------------------%
\section{Experiments}
\label{sec:experiments}
Next, we evaluated the baseline methods shown in Table \ref{tab:baselines}. As mentioned above, we did not consider the recently proposed neutral baseline \cite{izzo2020baseline} in this evaluation. All experiments were conducted on an NVIDIA GeForce GTX 1050 TI GPU with Intel i5 7500 CPU and 16Gb RAM. Our machine ran Linux Fedora 32 and Python 3.7.6.

We selected four state-of-the-art local attribution methods to illustrate the effect of different baselines. Specifically, we used KernelSHAP \cite{lundberg2017unified}, DeepSHAP \cite{lundberg2017unified}, DeepLift \cite{shrikumar2017learning} and Integrated Gradients (IG) \cite{sundararajan2017axiomatic}. Note that the authors of SHAP, \citet{lundberg2017unified}, substitute missing features based on the training distribution, which we called expectation baseline above. \citet{lundberg2017unified} thereby aim to align their work with earlier methods that approximate the Shapley value. Still, it is worth considering the SHAP framework in our evaluation, as some applications may require different baselines. In contrast, DeepLift and IG leave the choice of a baseline to the user, apart from suggesting the zero vector as a meaningful baseline for image recognition tasks. Note that all open source packages readily allow the user to set a baseline.

We used a top-$K$ ablation test to quantify the discriminative power of the generated feature attributions regarding the different baselines. Accordingly, we masked $K$ percent of the most highly attributed input features with random noise and measured the effect on the generated F1 score. The F1 score is the harmonic mean of precision and recall and provides valid results, even if the target class is imbalanced. Ablation tests are a popular and intuitive evaluation technique for feature attribution methods. Still, ablation tests should always be considered with care, since they do not consider feature interactions. In general, the evaluation of explanation models is subject to ongoing discussions in the research community.

For DeepSHAP, DeepLIFT and IG we trained a simple neural network with one hidden layer and a ReLu activation, using a sigmoid activation function at the output layer. Note that we deliberately chose a shallow architecture, since complex nets tend to mitigate noise at the input. In this way, we wanted to maintain the effect that masking certain input features had on predictive performance, thus providing an unbiased view of the effect of the different baselines. Finally, since KernelSHAP is model agnostic, we applied it to a Support Vector Machine.

As discussed above, the blurred baseline assumes some form of neighbourhood among input features. Since tabular data is usually ordered arbitrarily, we computed the blurred baseline for 1,000 feature permutations and averaged the results. Accordingly, we examined every feature with respect to different neighbouring features, thereby mitigating the effect of the initial ordering of the features.

All experiments are also available on our GitHub page.\footnote{https://github.com/ITZuern/On-Baselines-for-Local-Feature-Attributions}

%---------- Complete Ablation Tests --------------------------------%
\begin{figure*}[htp]
\centering
\subfigure[DeepLift \cite{shrikumar2017learning} on Human Activity Recognition]{
    \includegraphics[width=0.23\textwidth]{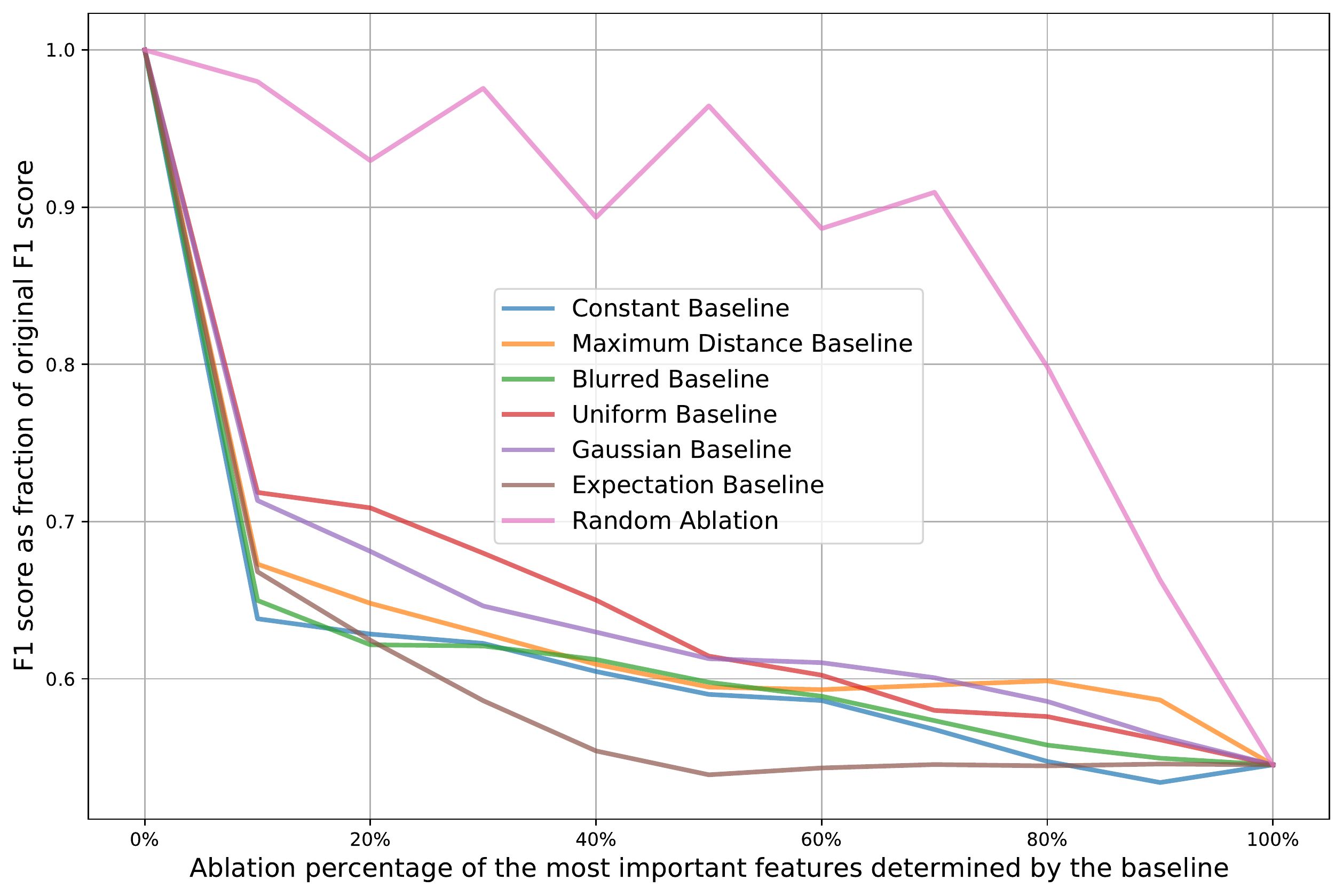}
}
\hfill
\subfigure[IG \cite{sundararajan2017axiomatic} on Human Activity Recognition]{
    \includegraphics[width=0.23\textwidth]{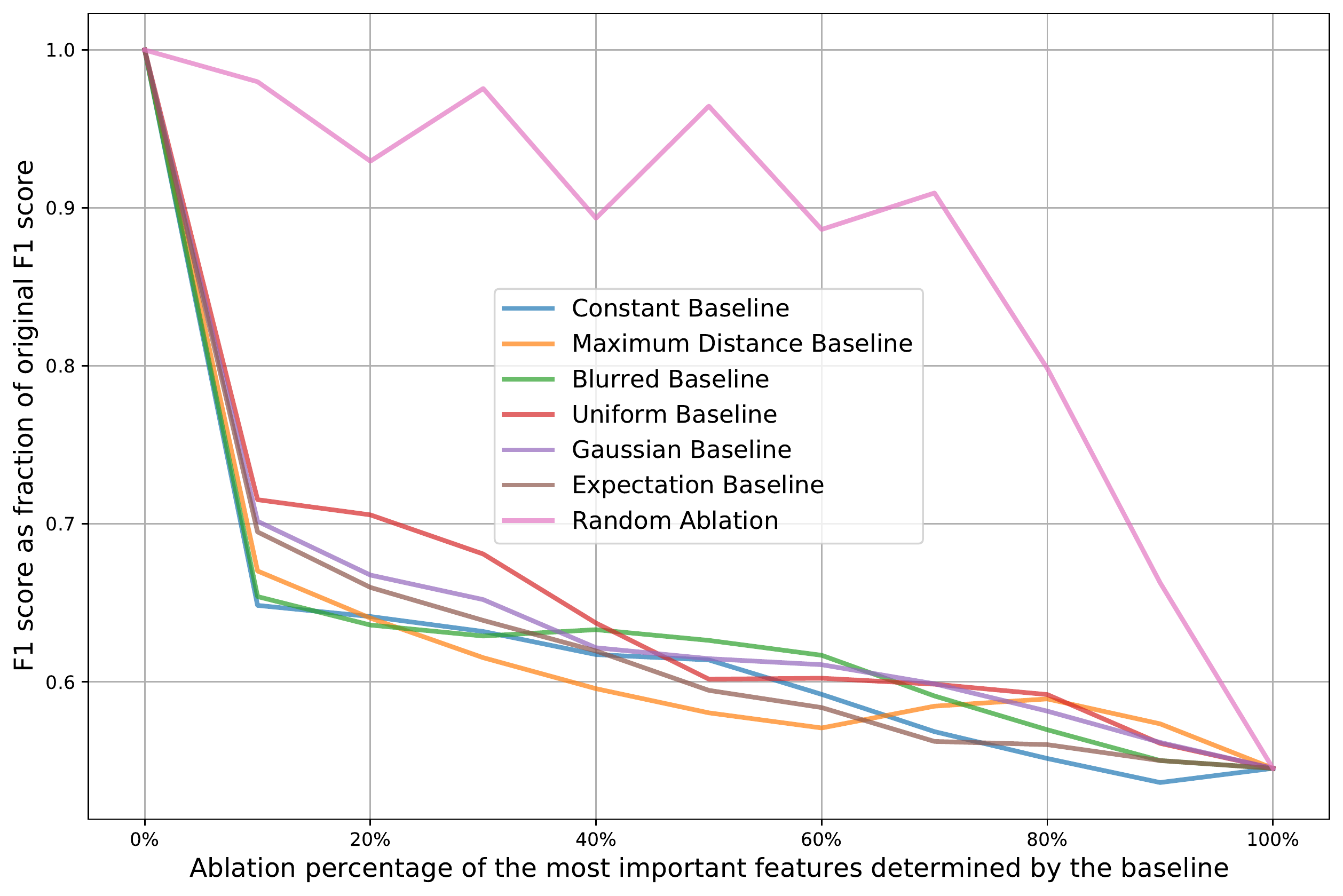}
}
\hfill
\subfigure[KernelSHAP \cite{lundberg2017unified} on Human Activity Recognition]{
    \includegraphics[width=0.23\textwidth]{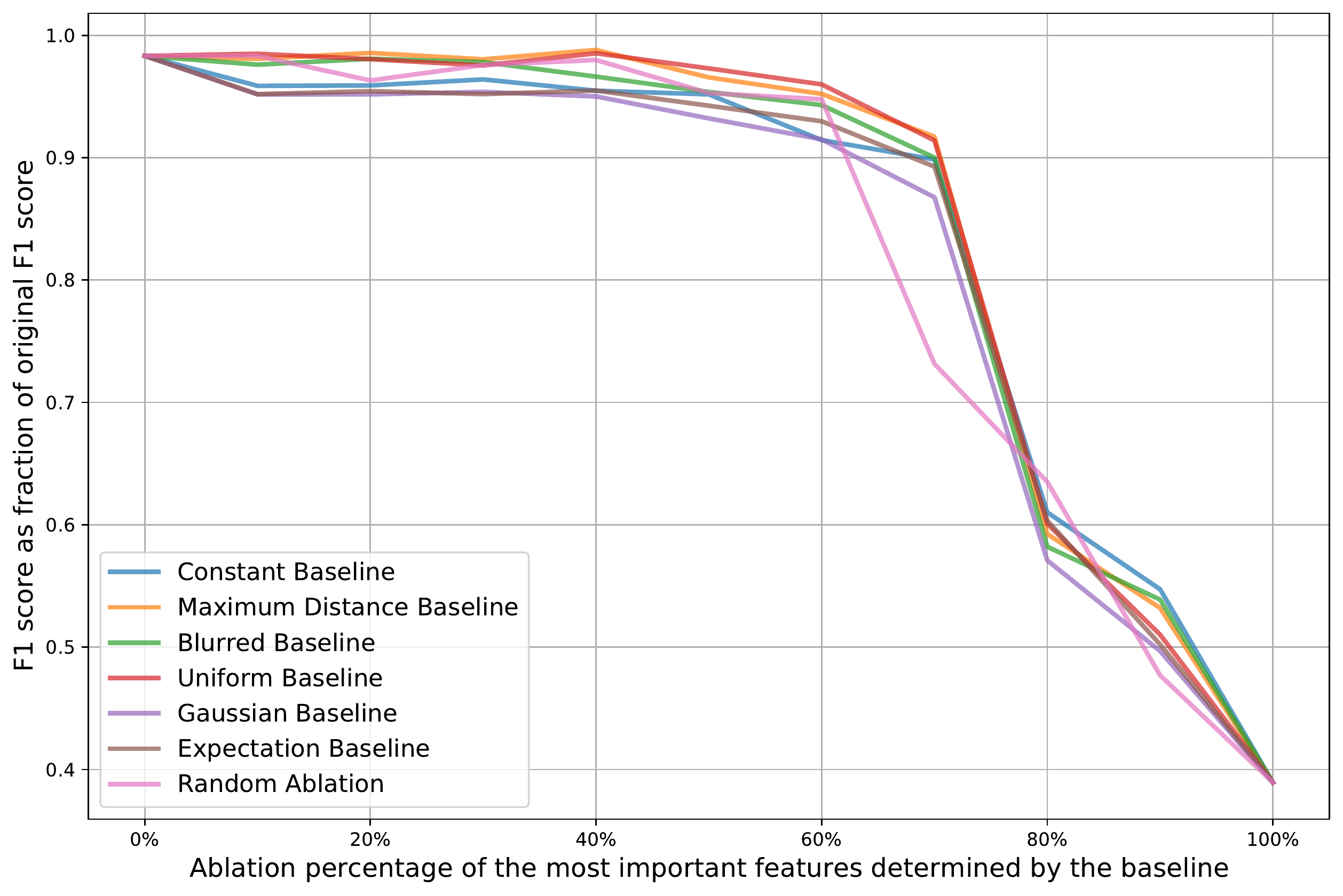}
}
\hfill
\subfigure[DeepSHAP \cite{lundberg2017unified} on Human Activity Recognition]{
    \includegraphics[width=0.23\textwidth]{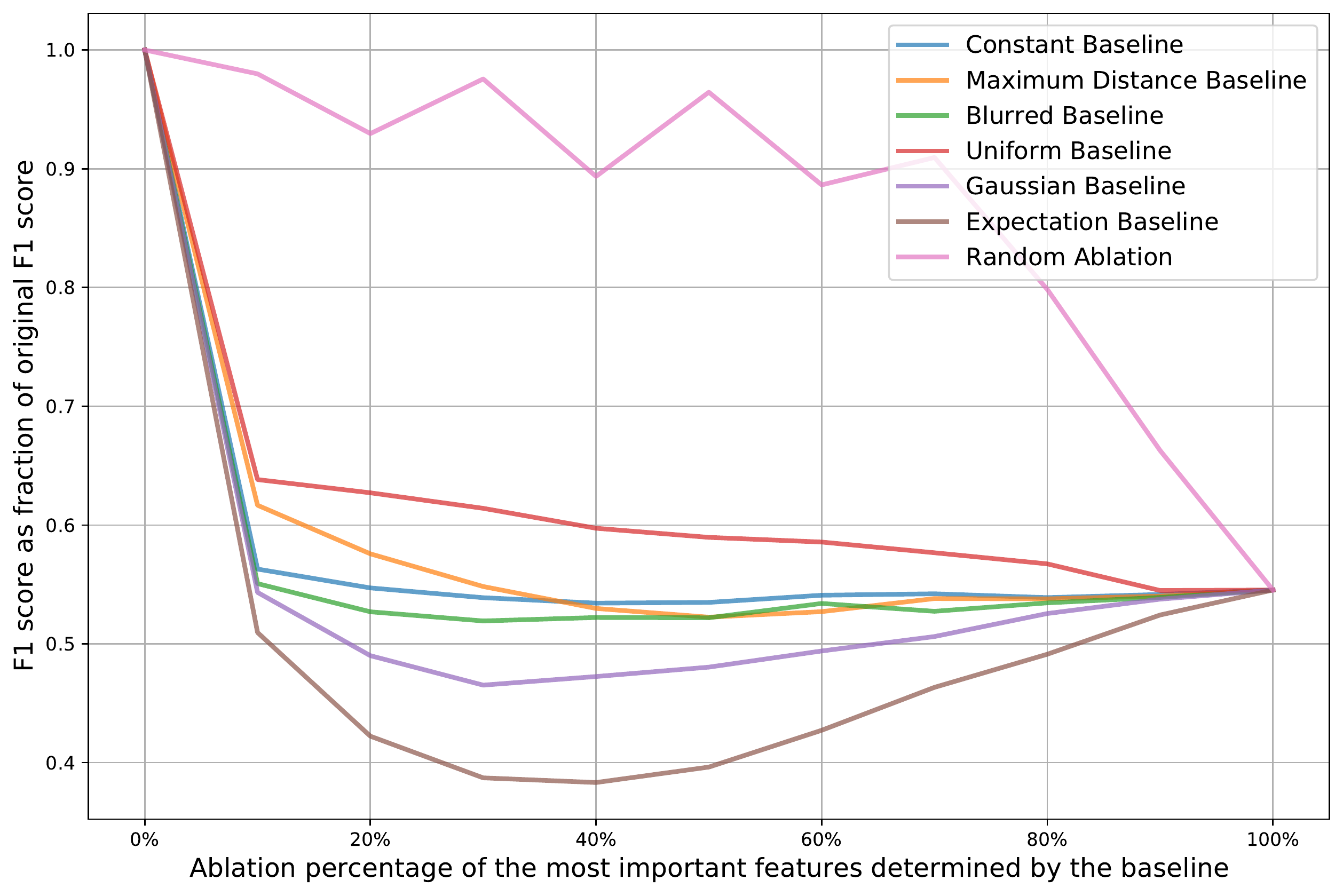}
}
\hfill
\subfigure[DeepLift \cite{shrikumar2017learning} on Fraud Detection]{
    \includegraphics[width=0.23\textwidth]{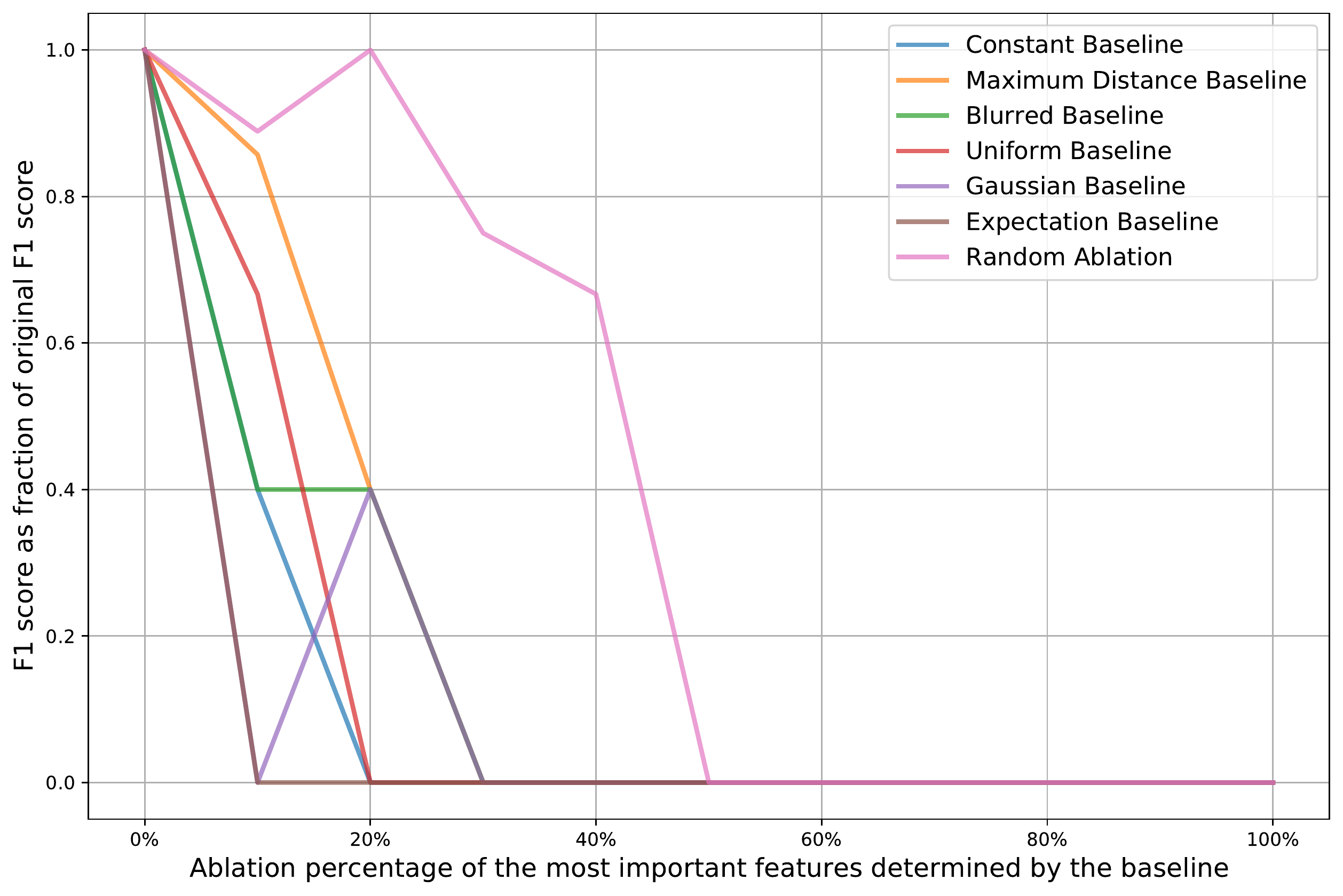}
}
\hfill
\subfigure[IG \cite{sundararajan2017axiomatic} on Fraud Detection]{
    \includegraphics[width=0.23\textwidth]{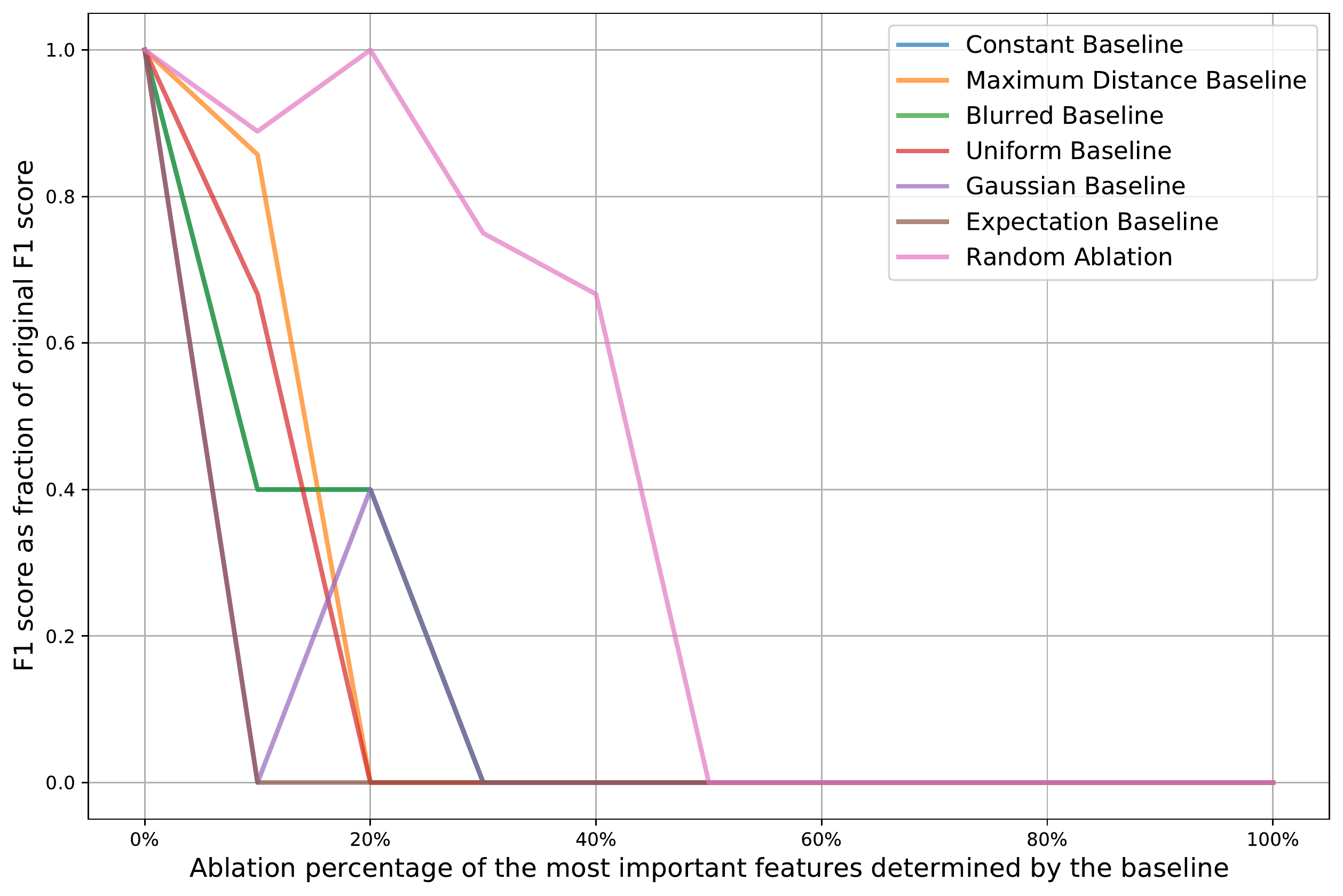}
}
\hfill
\subfigure[KernelSHAP \cite{lundberg2017unified} on Fraud Detection]{
    \includegraphics[width=0.23\textwidth]{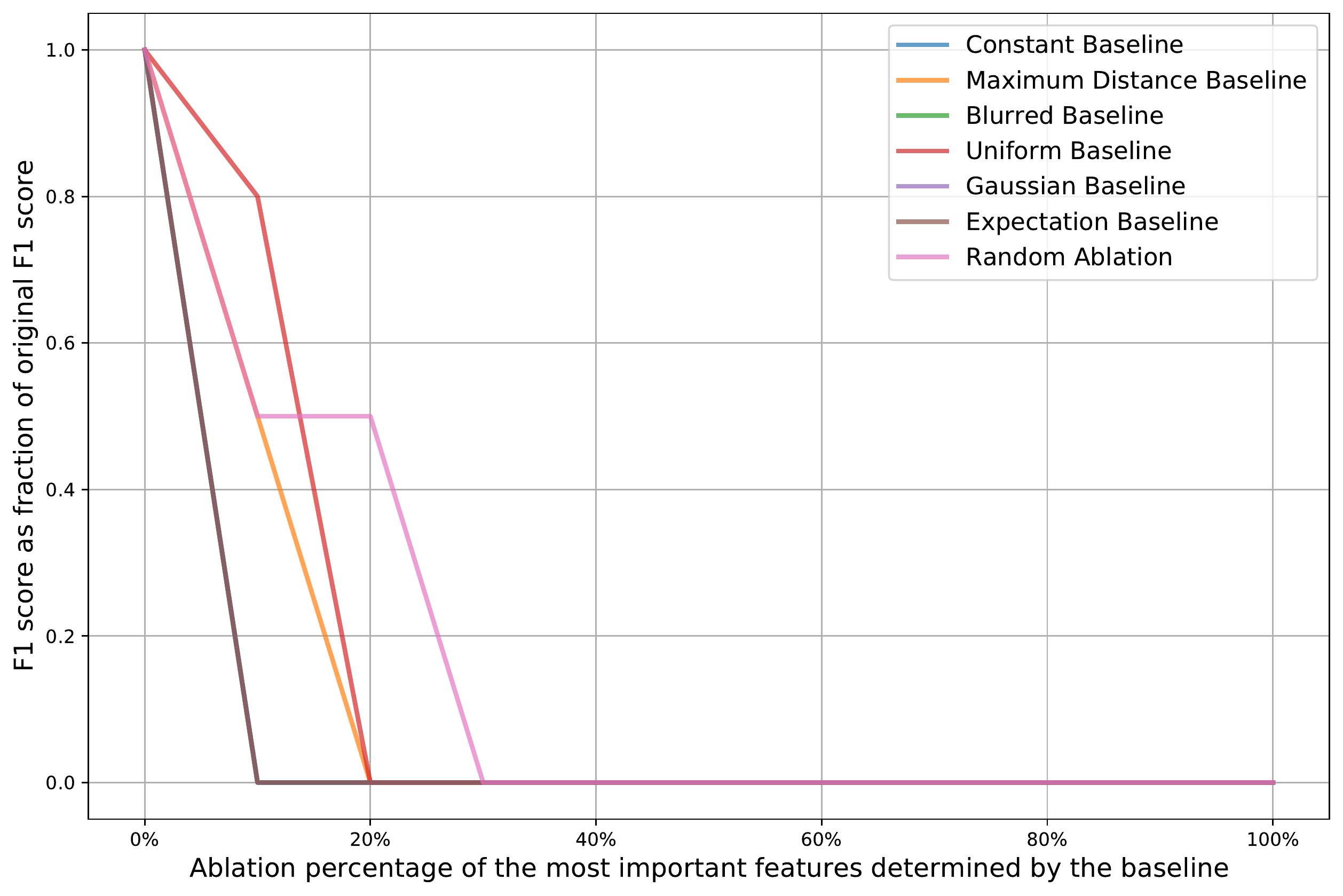}
}
\hfill
\subfigure[DeepSHAP \cite{lundberg2017unified} on Fraud Detection]{
    \includegraphics[width=0.23\textwidth]{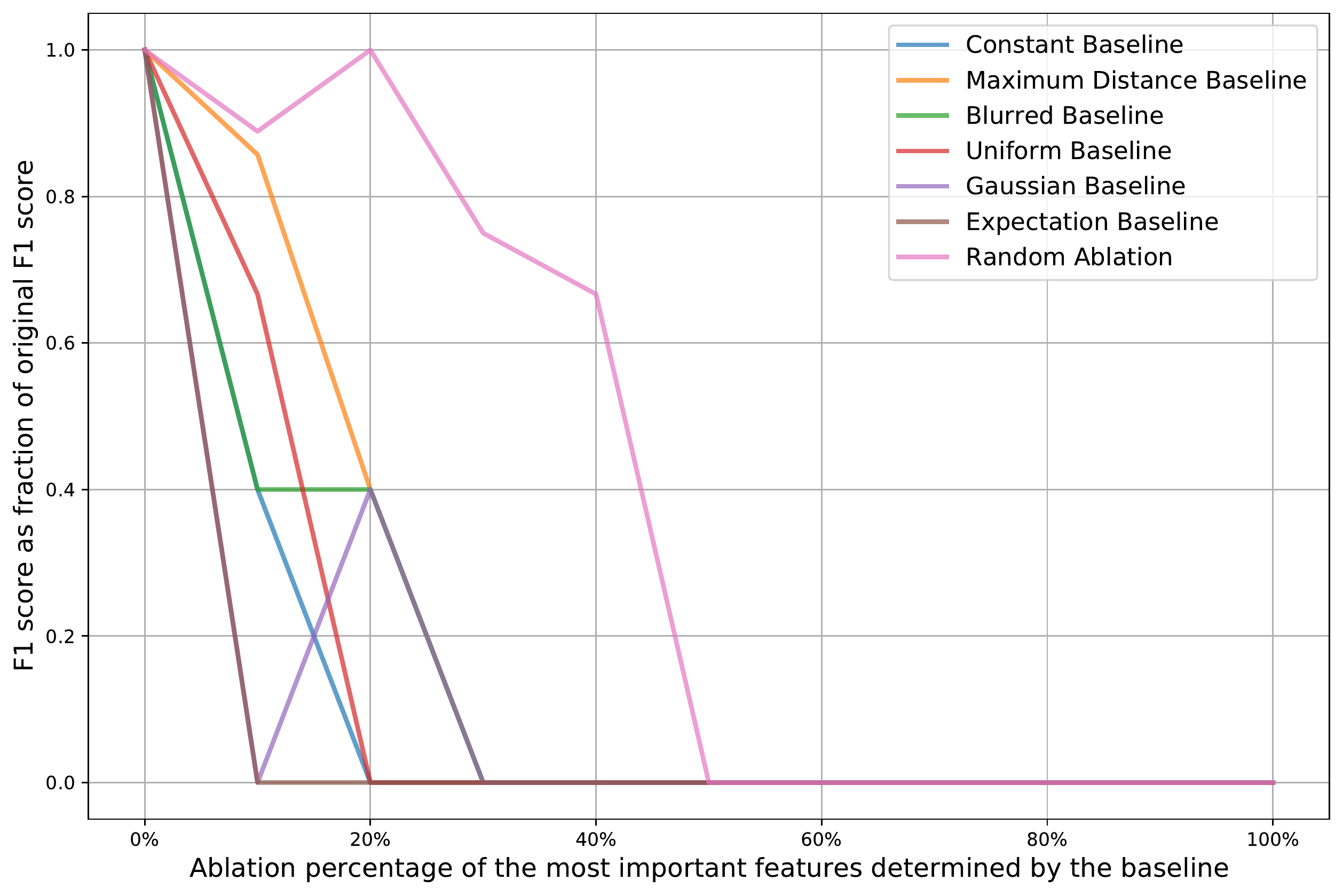}
}
\hfill
\subfigure[DeepLift \cite{shrikumar2017learning} on Communities]{
    \includegraphics[width=0.23\textwidth]{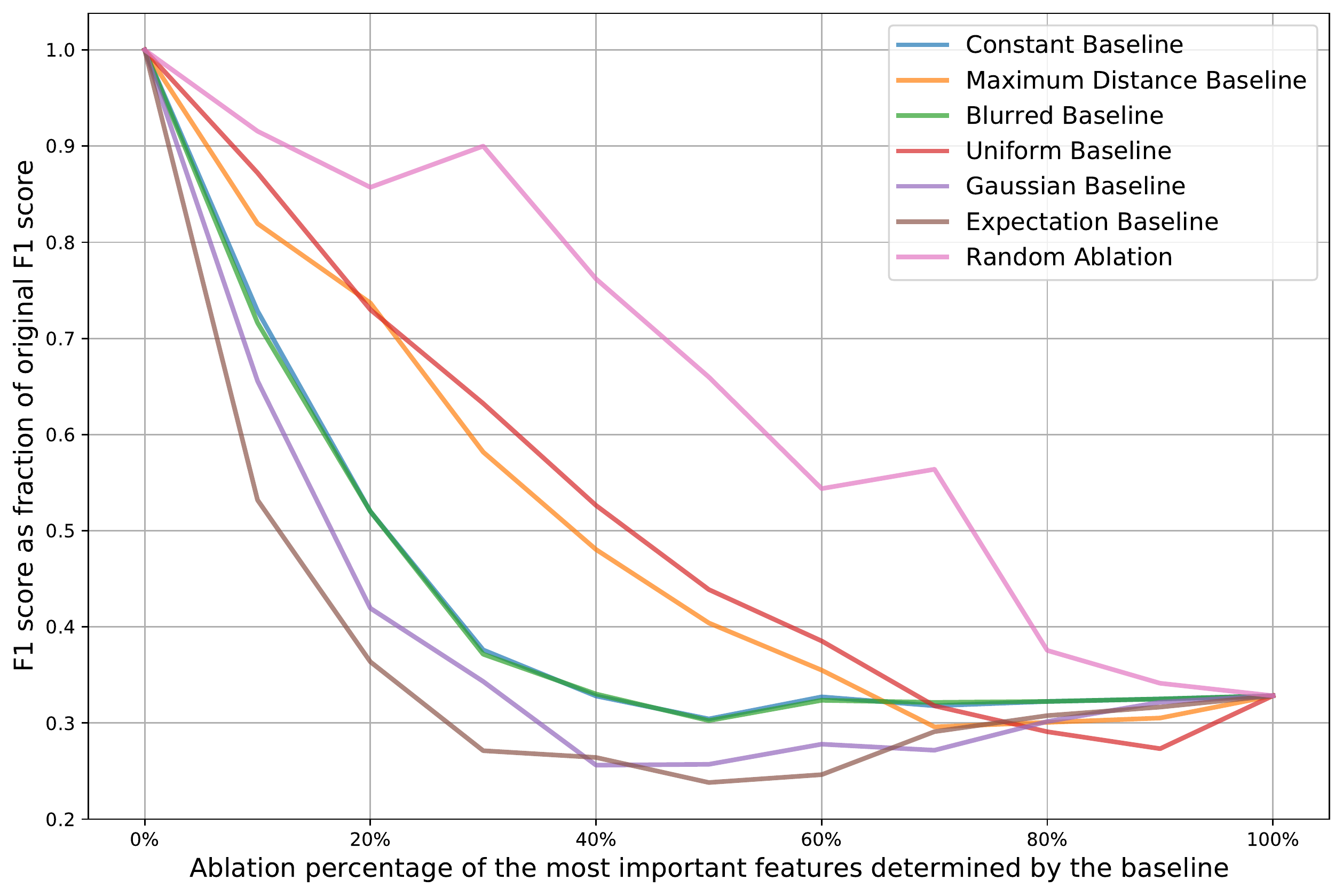}
}
\hfill
\subfigure[IG \cite{sundararajan2017axiomatic} on Communities]{
    \includegraphics[width=0.23\textwidth]{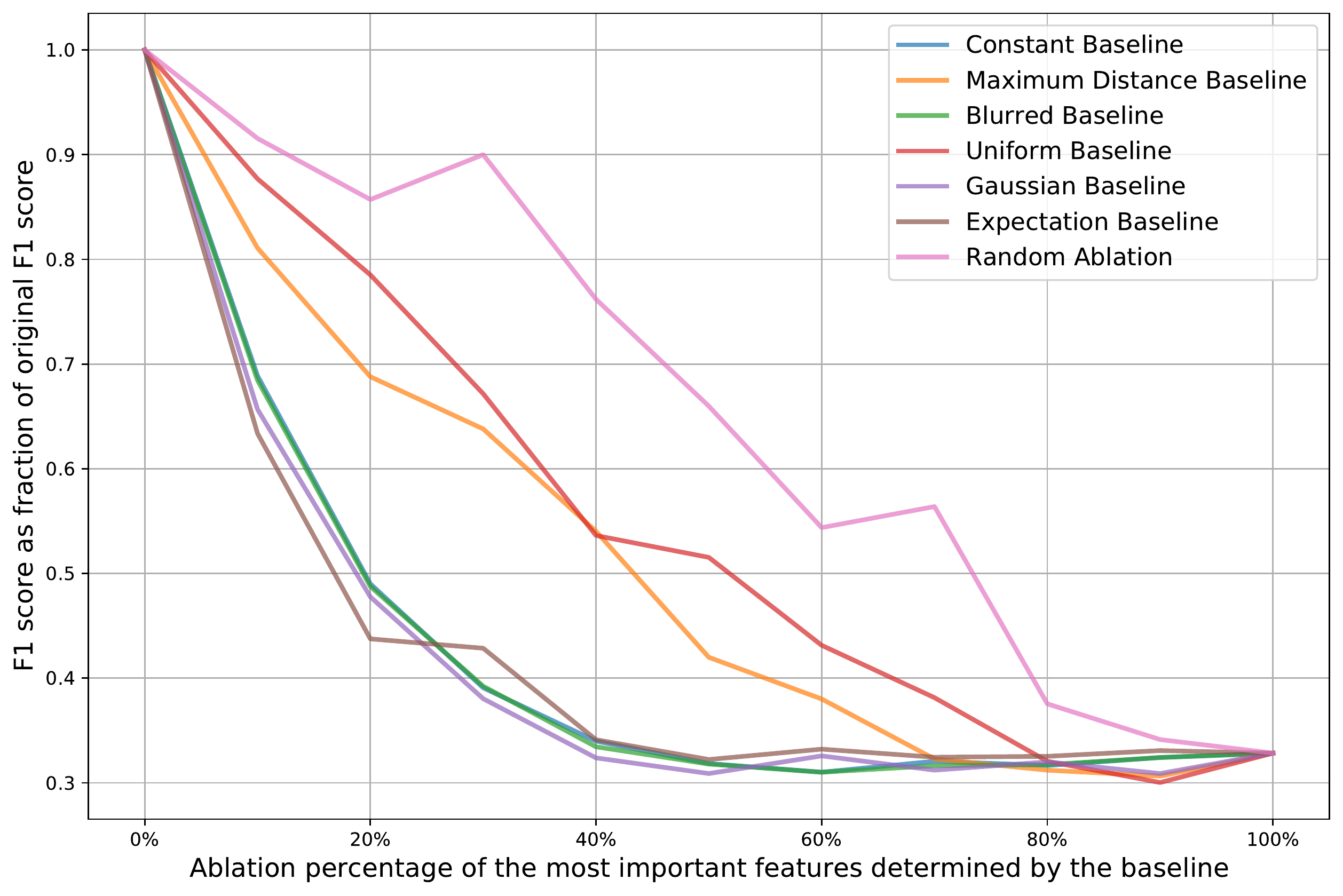}
}
\hfill
\subfigure[KernelSHAP \cite{lundberg2017unified} on Communities]{
    \includegraphics[width=0.23\textwidth]{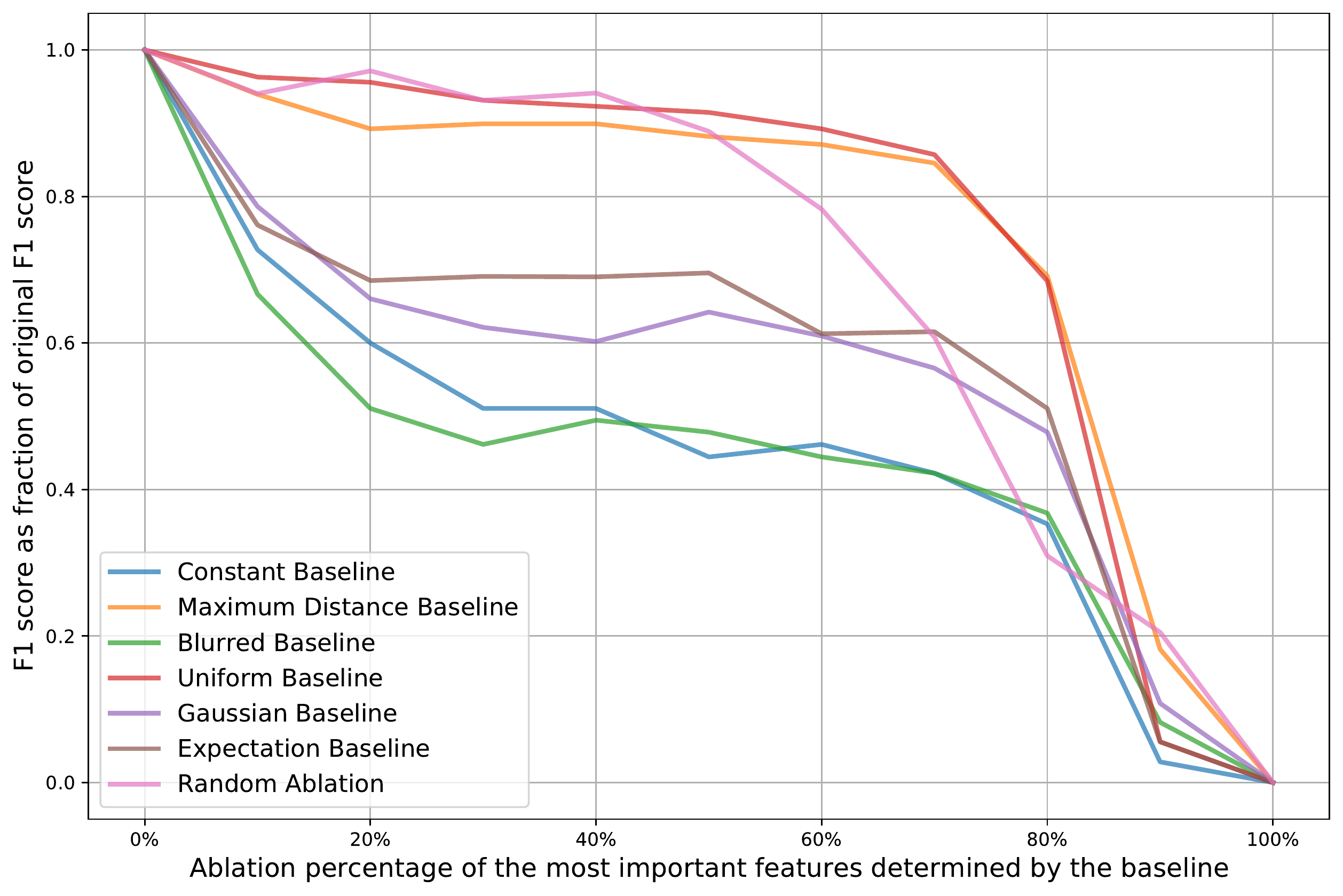}
}
\hfill
\subfigure[DeepSHAP \cite{lundberg2017unified} on Communities]{
    \includegraphics[width=0.23\textwidth]{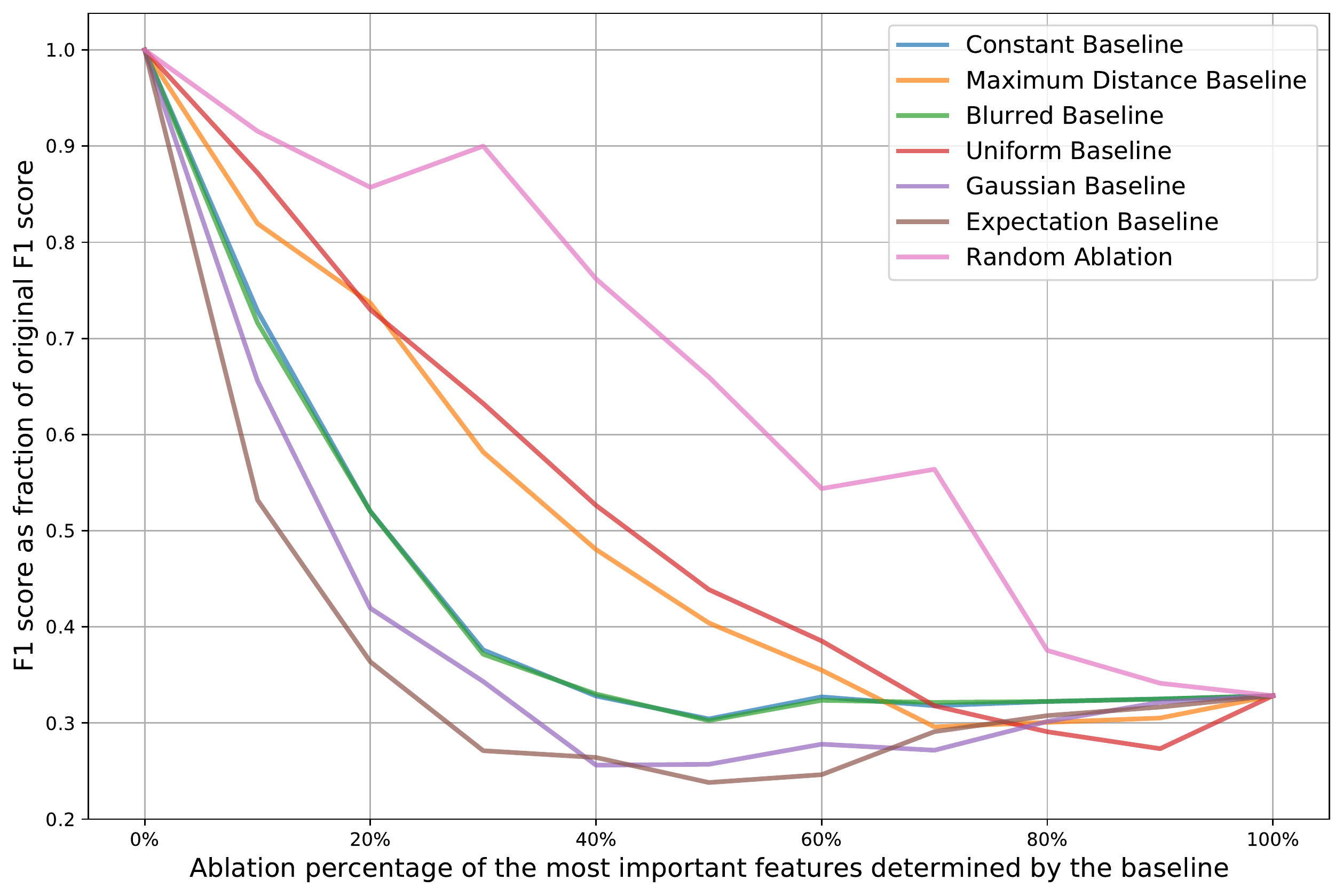}
}
\hfill
\subfigure[DeepLift \cite{shrikumar2017learning} on Spambase]{
    \includegraphics[width=0.23\textwidth]{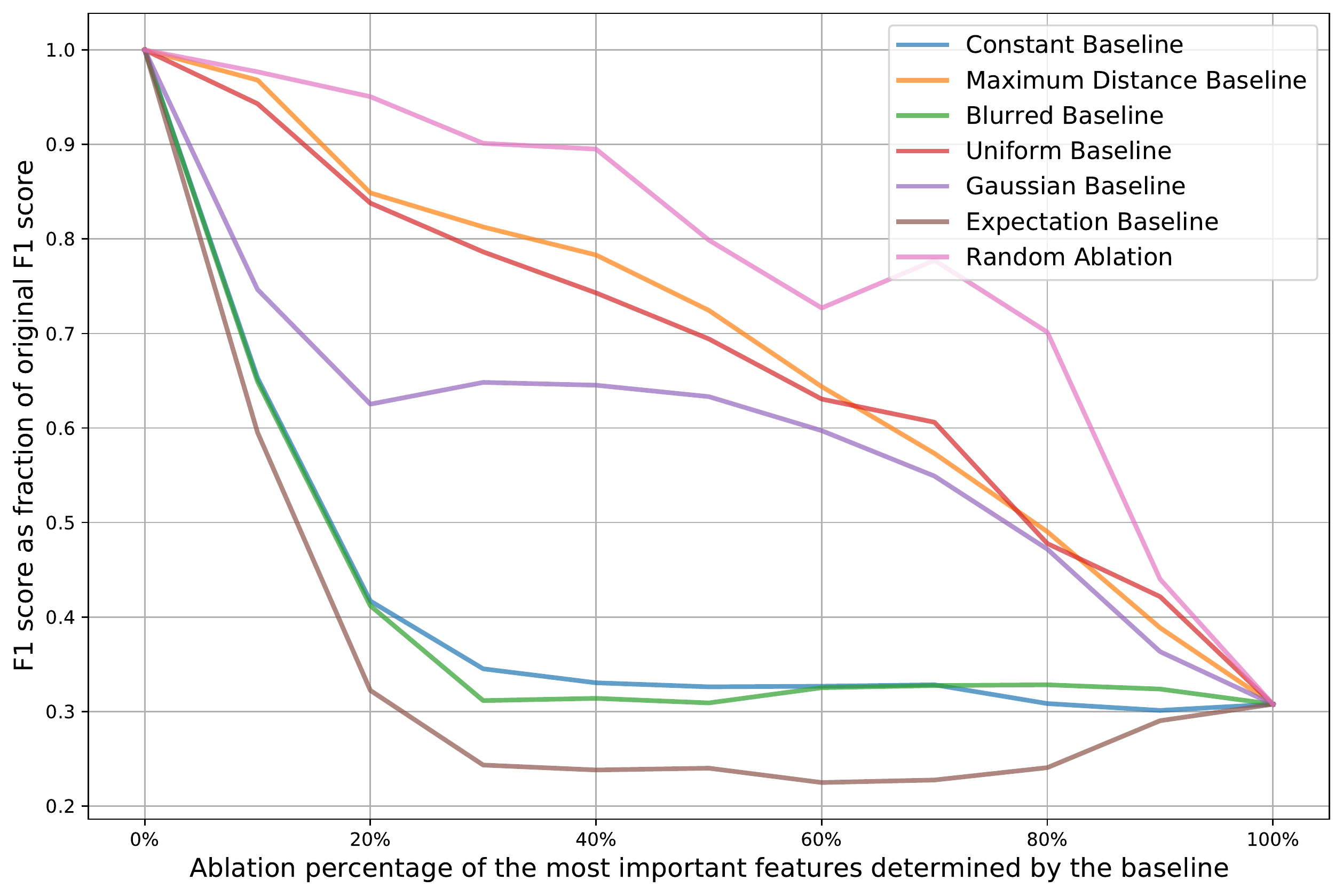}
}
\hfill
\subfigure[IG \cite{sundararajan2017axiomatic} on Spambase]{
    \includegraphics[width=0.23\textwidth]{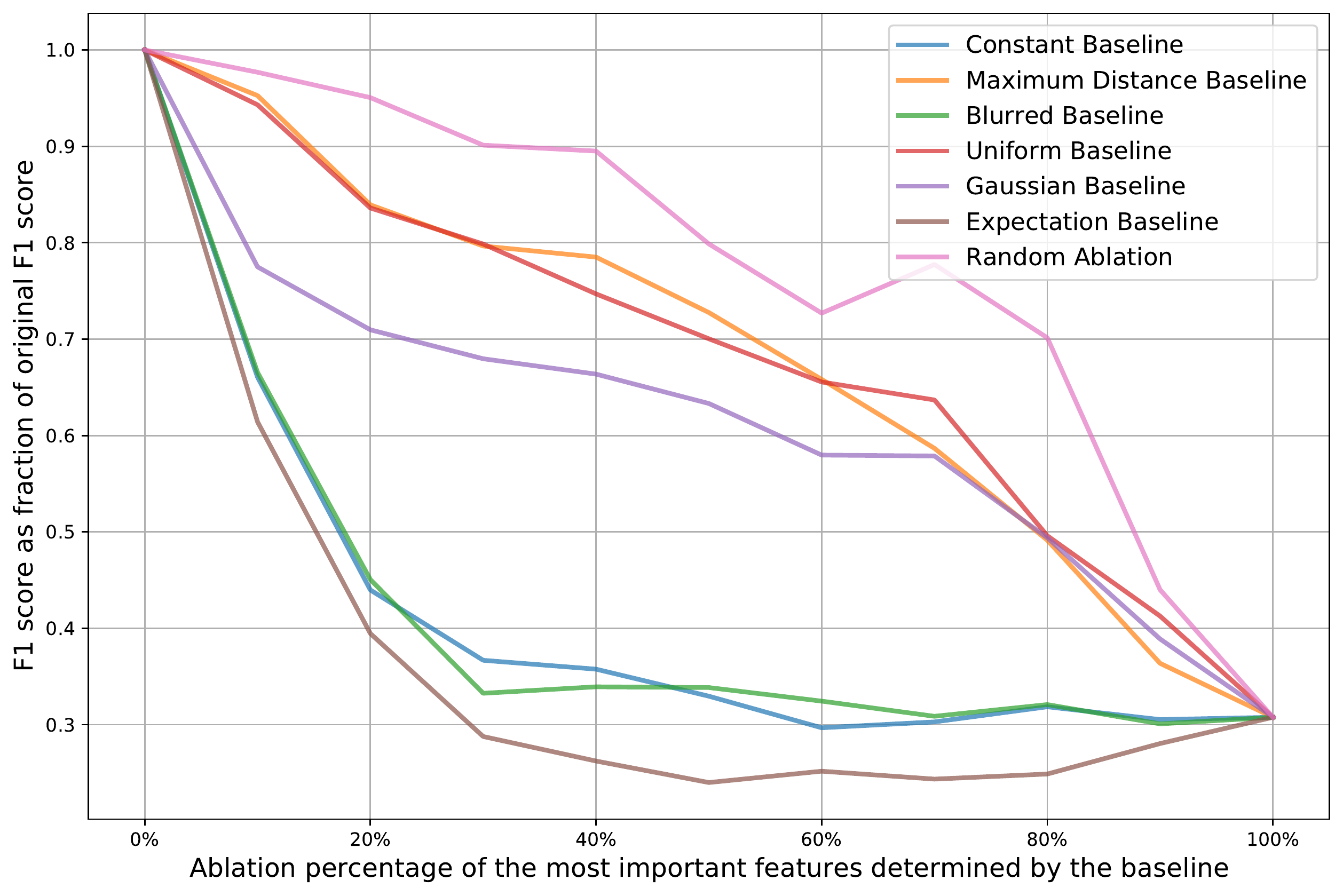}
}
\hfill
\subfigure[KernelSHAP \cite{lundberg2017unified} on Spambase]{
    \includegraphics[width=0.23\textwidth]{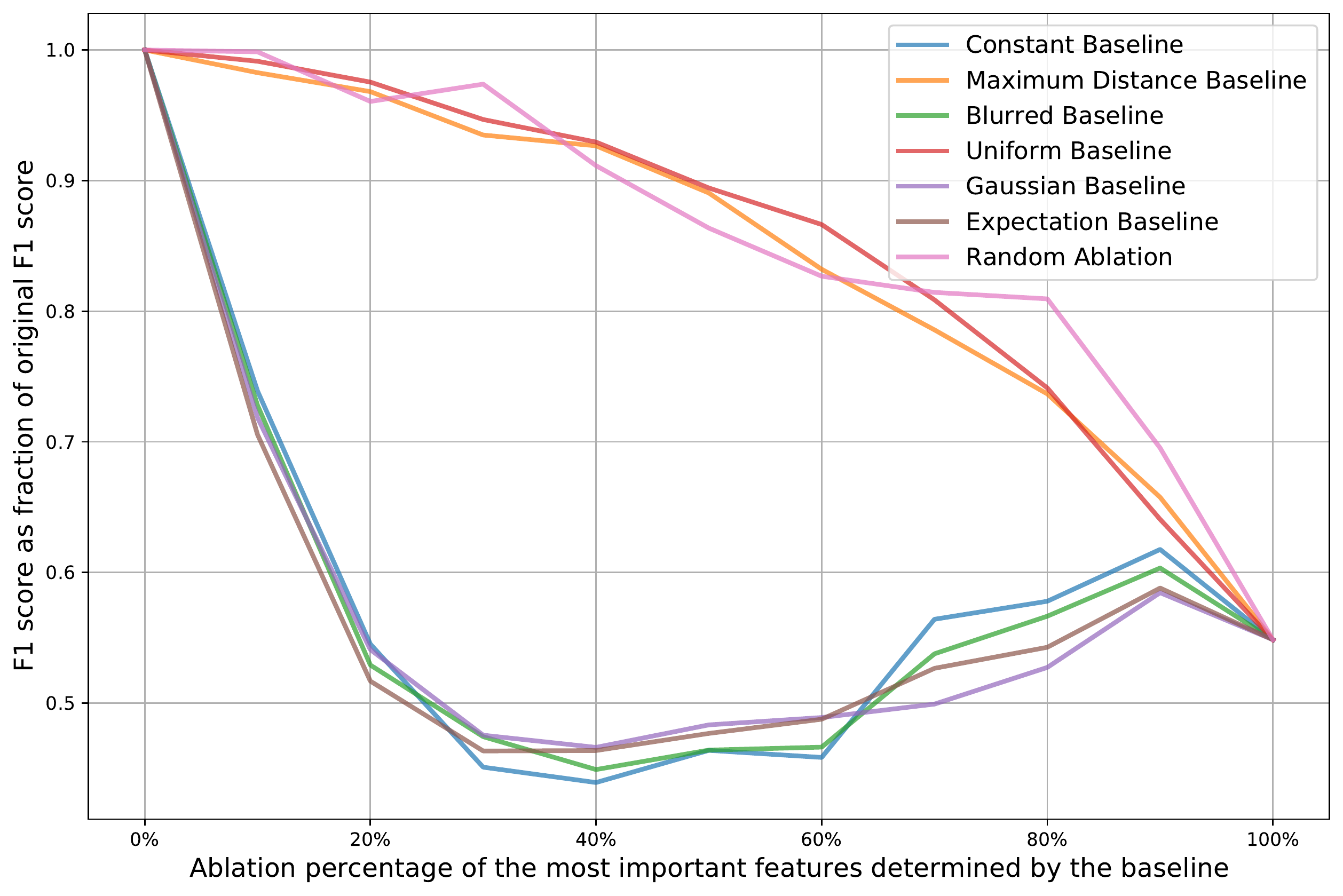}
}
\hfill
\subfigure[DeepSHAP \cite{lundberg2017unified} on Spambase]{
    \includegraphics[width=0.23\textwidth]{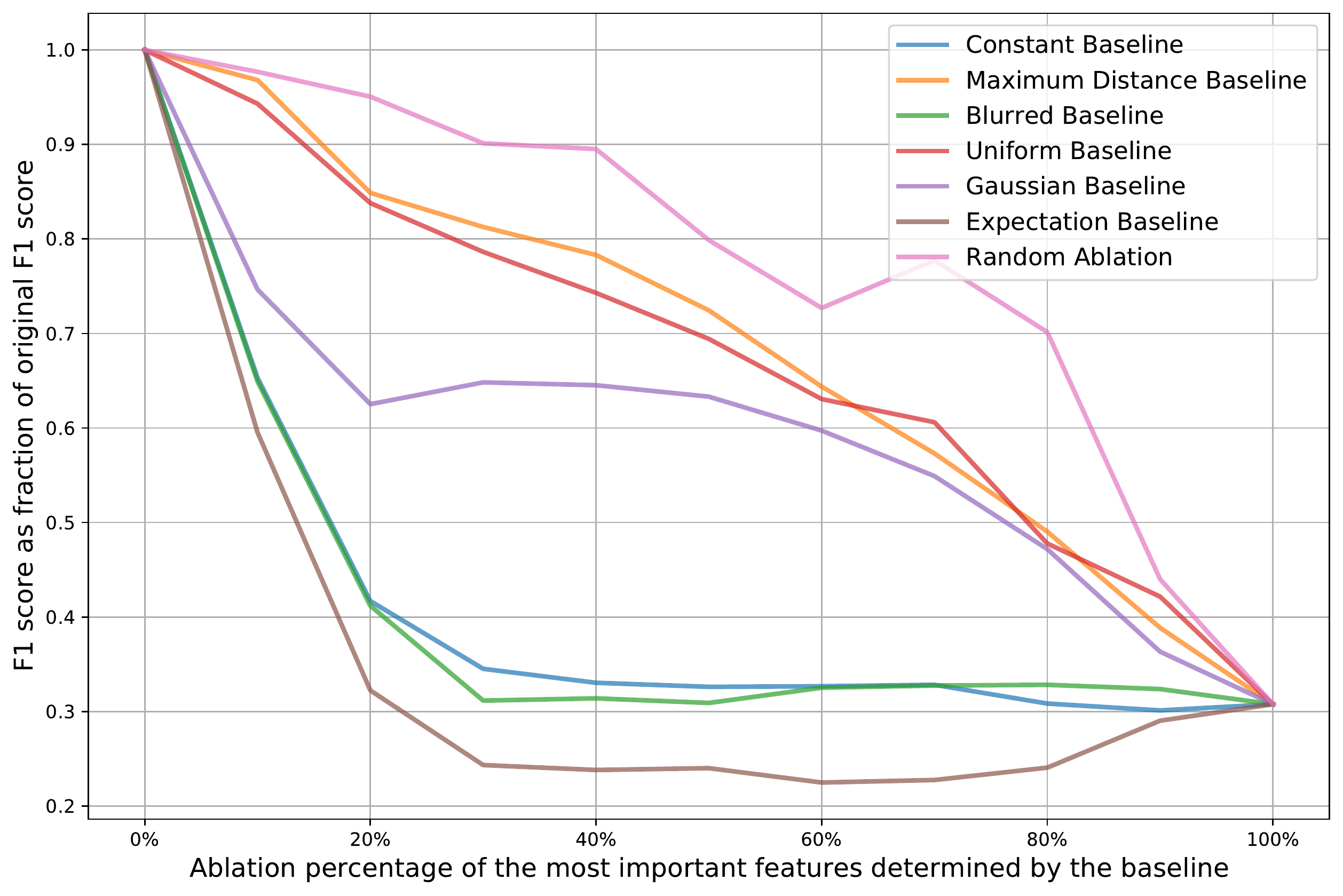}
}
\hfill
\subfigure[DeepLift \cite{shrikumar2017learning} on Compas]{
    \includegraphics[width=0.23\textwidth]{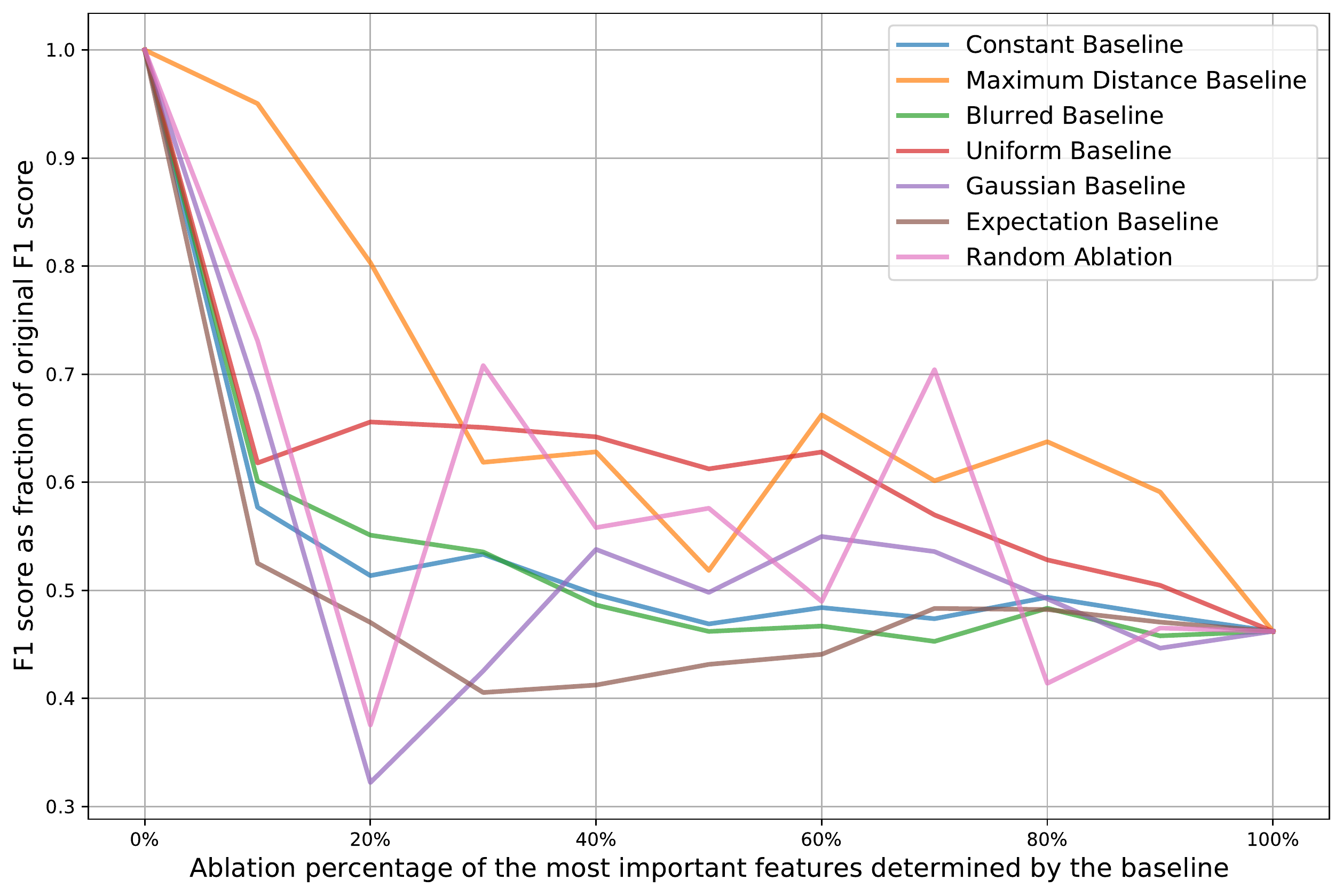}
}
\hfill
\subfigure[IG \cite{sundararajan2017axiomatic} on Compas]{
    \includegraphics[width=0.23\textwidth]{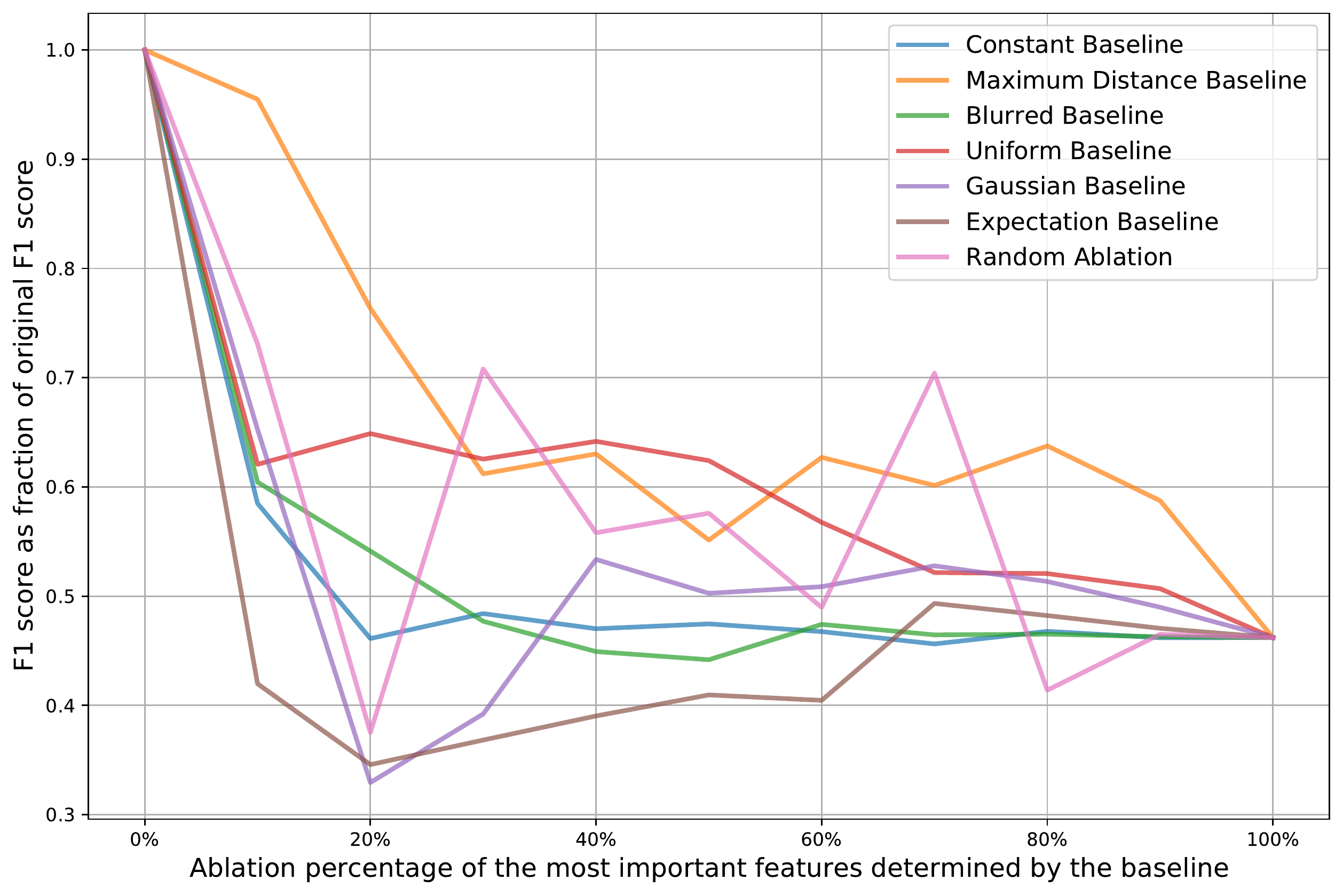}
}
\hfill
\subfigure[KernelSHAP \cite{lundberg2017unified} on Compas]{
    \includegraphics[width=0.23\textwidth]{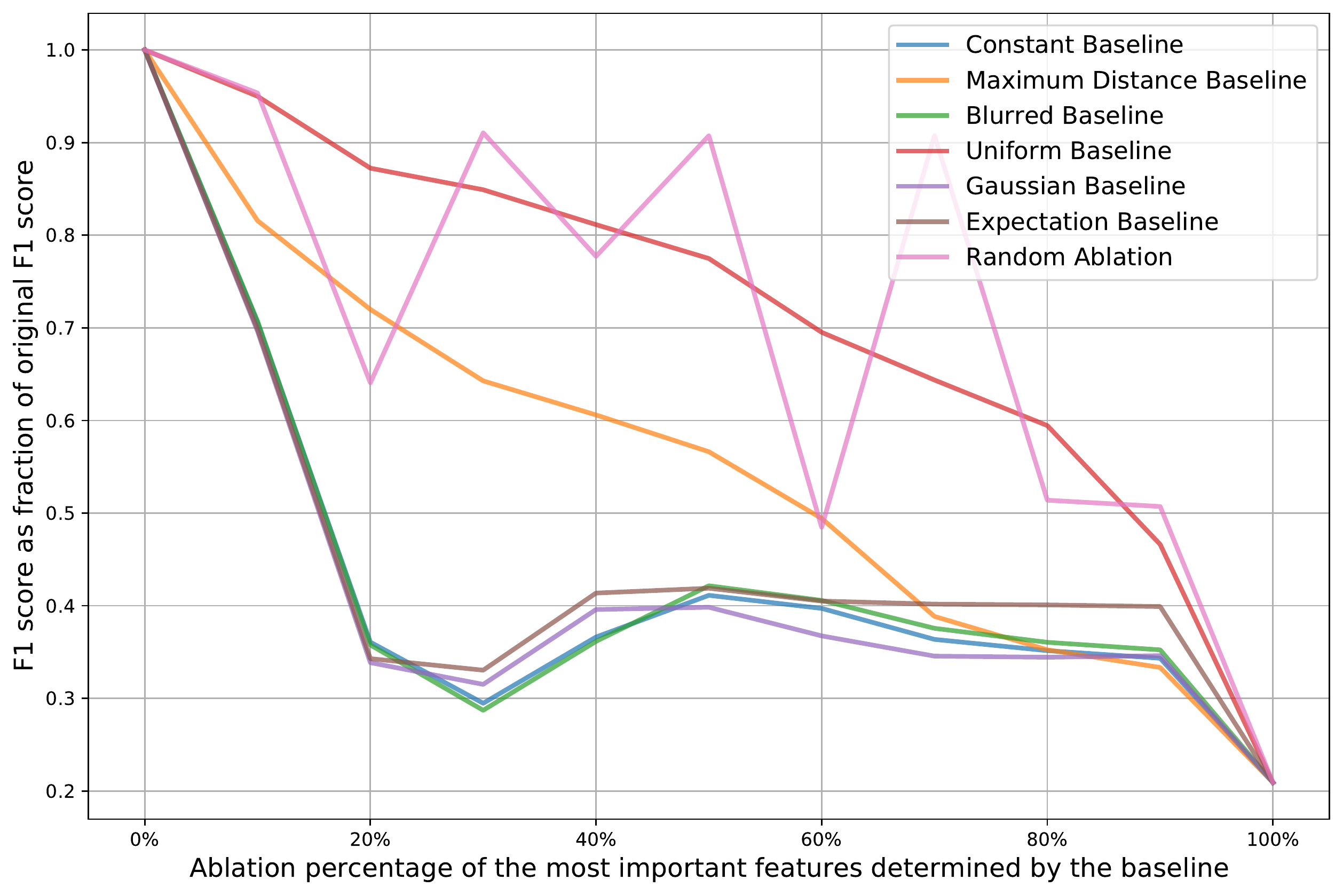}
}
\hfill
\subfigure[DeepSHAP \cite{lundberg2017unified} on Compas]{
    \includegraphics[width=0.23\textwidth]{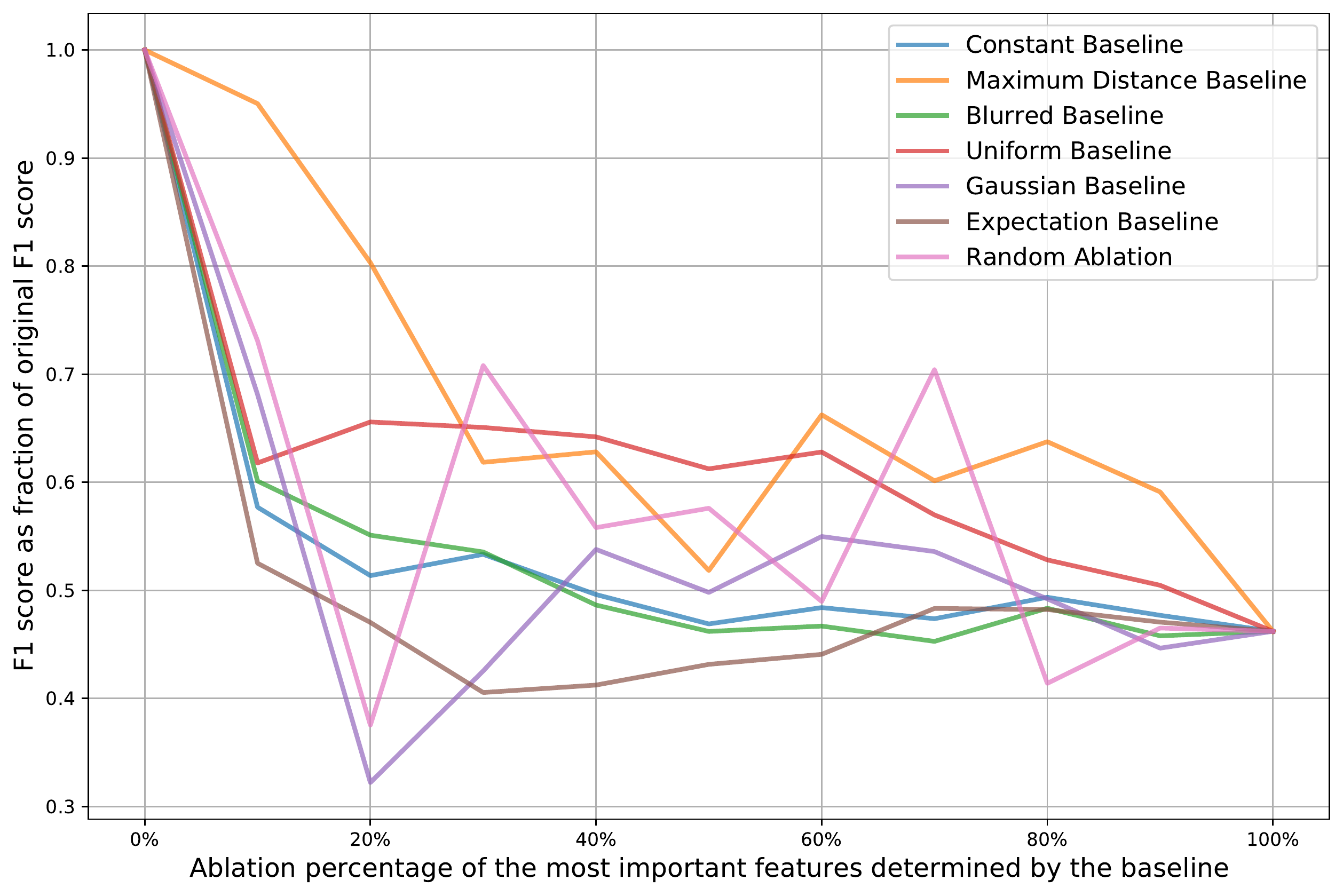}
}
\caption{\textbf{Top K ablation tests:} We conducted ablation tests for every data set, attribution model and baseline method (subplots). One might have to zoom into the subplots. Specifically, we masked $K$ percent (x-axis) of the most highly attributed features with random noise and monitored the change in the F1 score (y-axis). A large decrease in F1 indicates discriminative feature attributions with respect to the predictive model. Notably, there are significant differences between the baseline methods.}
\label{fig:ablation_per_dataset}
\end{figure*}
%---------- Complete Ablation Tests --------------------------------%
%---------- Average Ablation --------------------------------%
\begin{figure*}[ht]
\centering
\hspace*{\fill}
\subfigure[DeepLift \cite{shrikumar2017learning} on average over all data sets]{
    \includegraphics[width=0.34\textwidth]{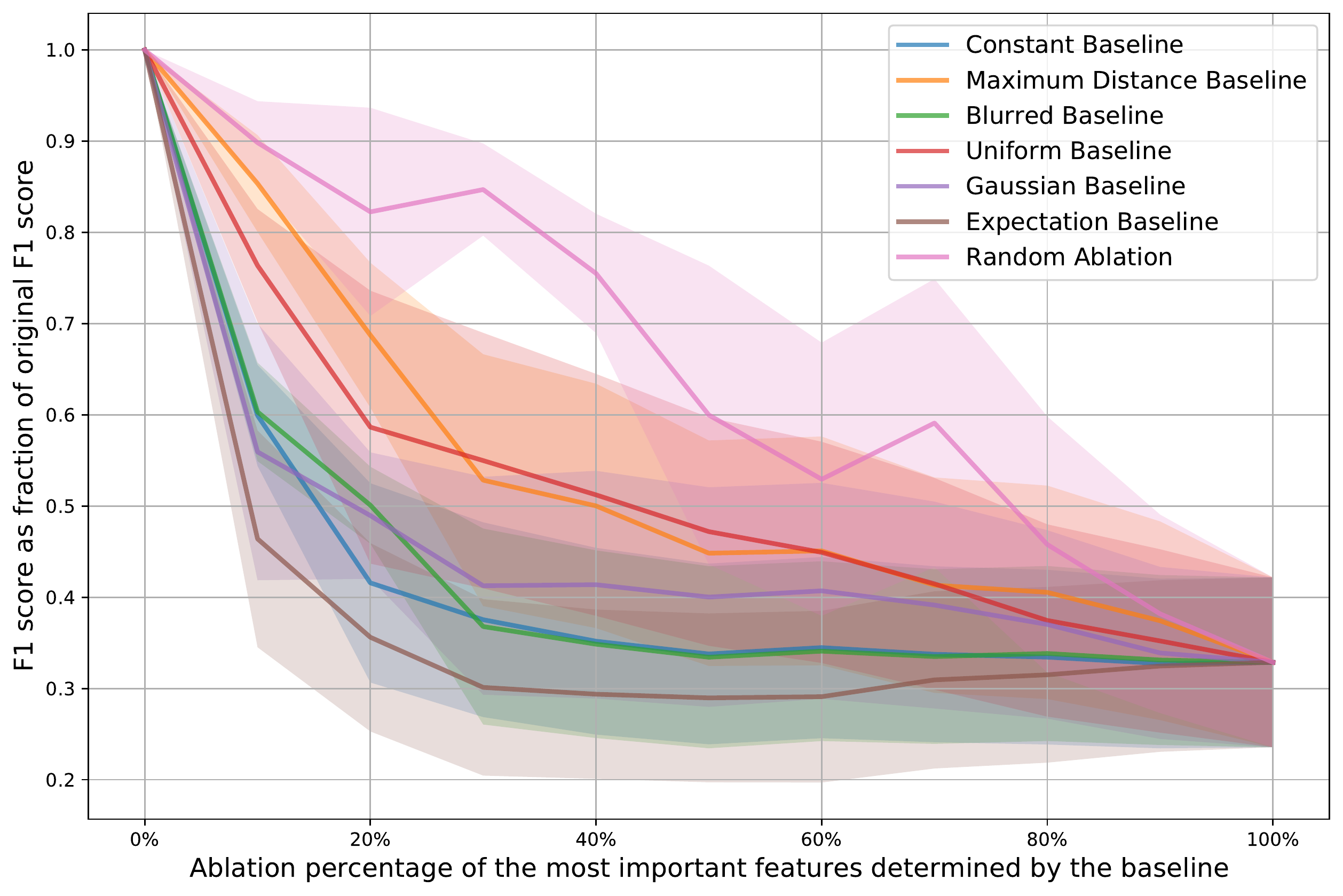}
}
\hfill
\subfigure[IG \cite{sundararajan2017axiomatic} on average over all data sets]{
    \includegraphics[width=0.34\textwidth]{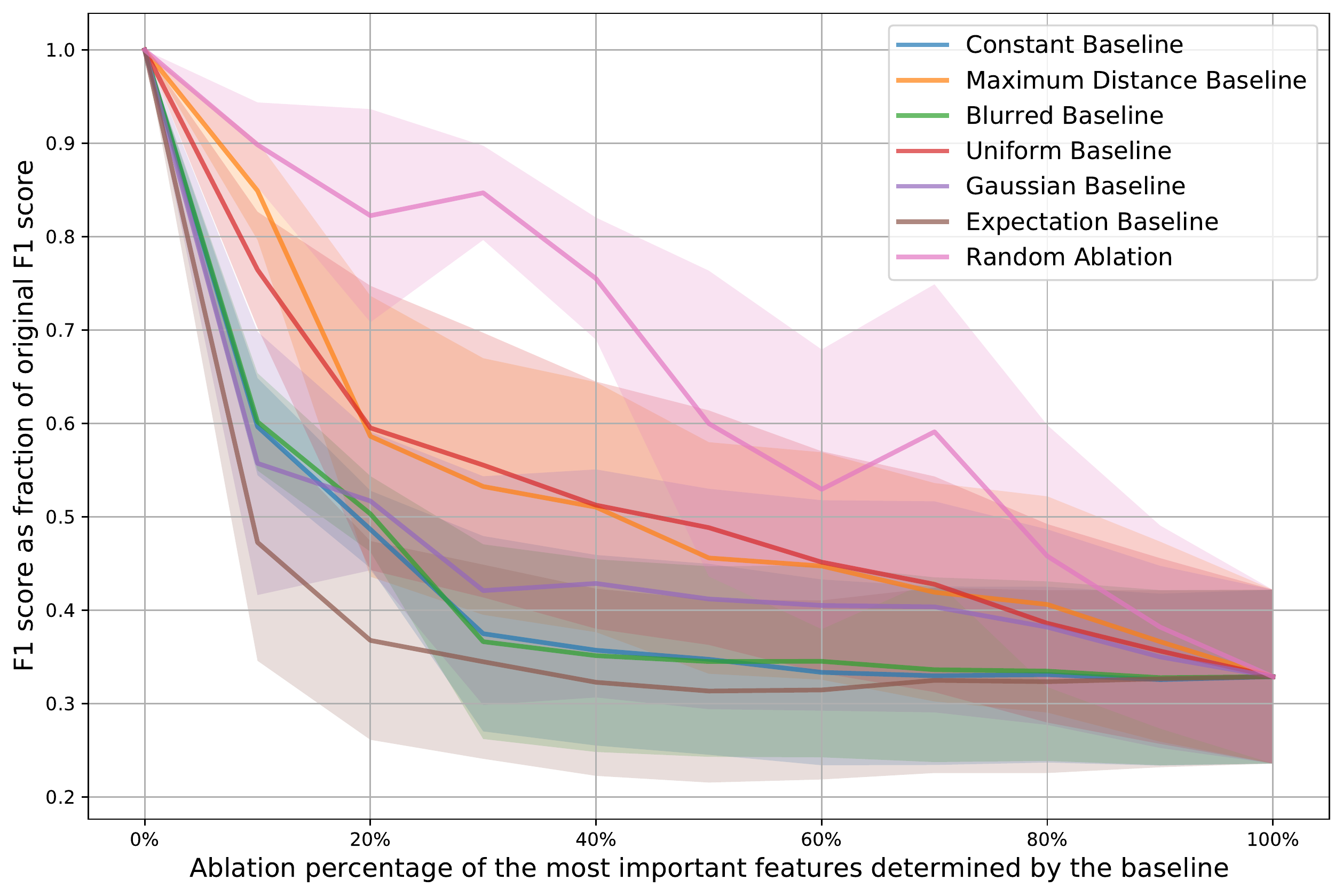}
}
\hspace*{\fill}

\hspace*{\fill}
\subfigure[KernelSHAP \cite{lundberg2017unified} on average over all data sets]{
    \includegraphics[width=0.34\textwidth]{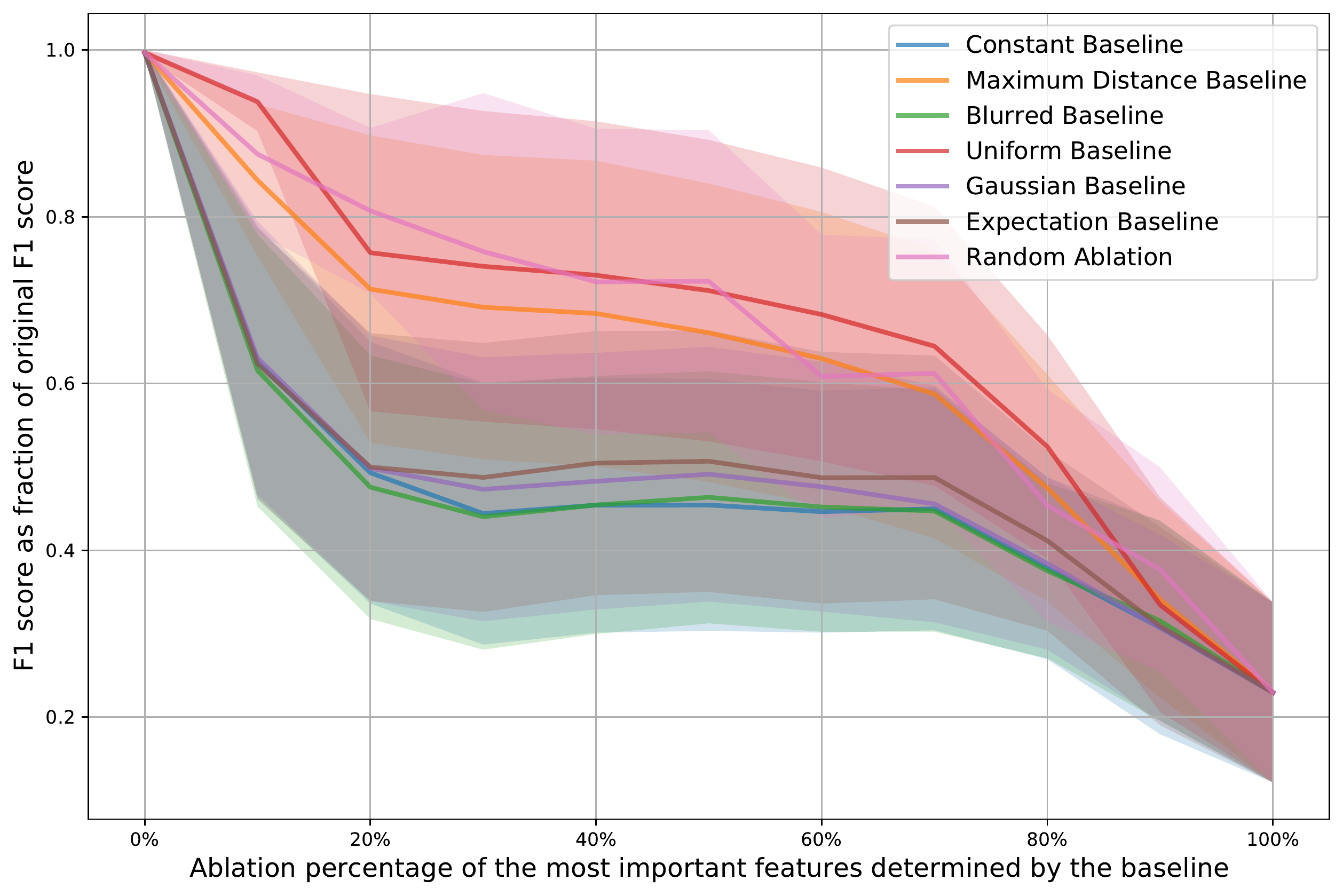}
}
\hfill
\subfigure[DeepSHAP \cite{lundberg2017unified} on average over all data sets]{
    \includegraphics[width=0.34\textwidth]{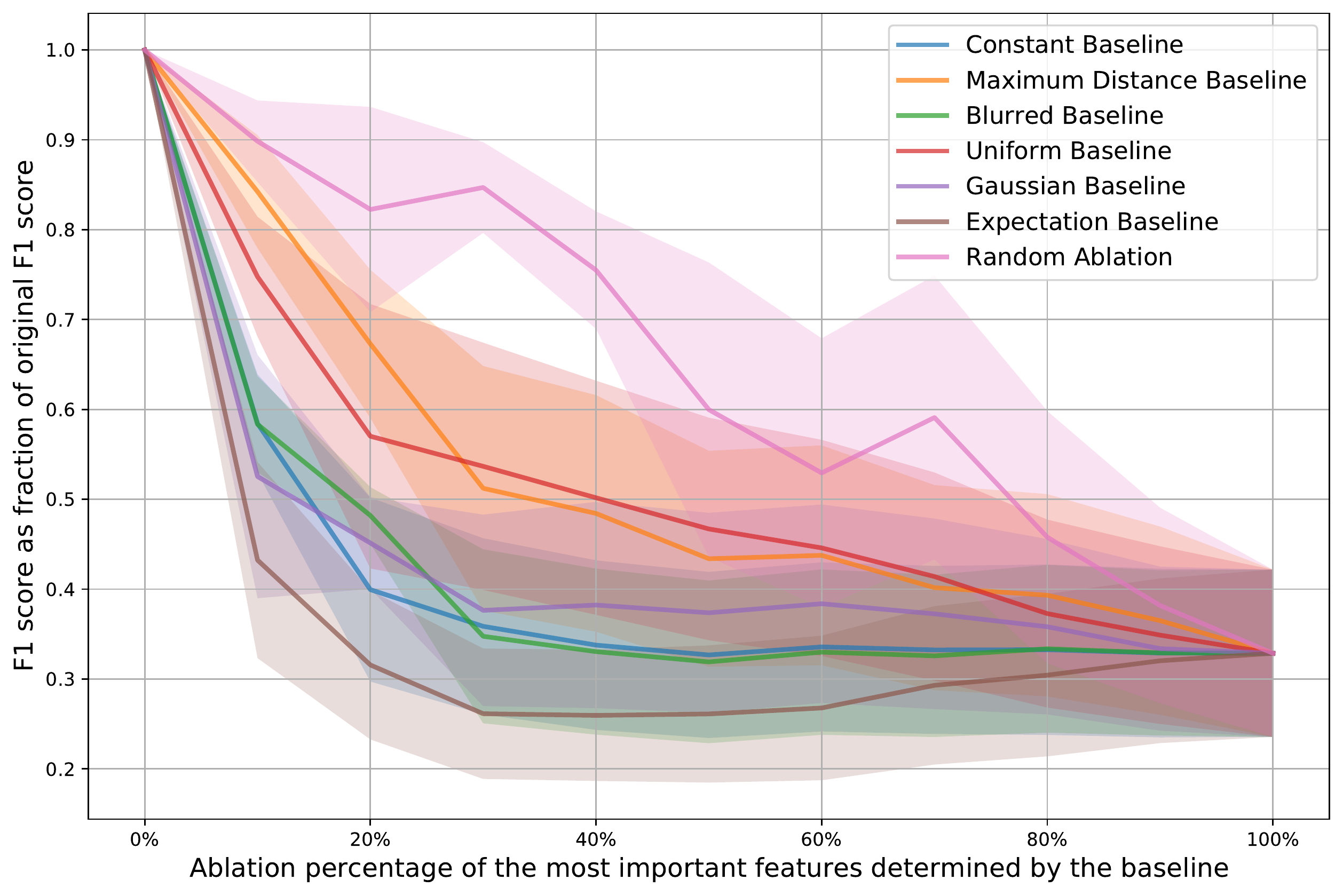}
}
\hspace*{\fill}
\caption{\textbf{Average ablation tests:} Here we depict the ablation test results averaged over all tabular data sets. The lines correspond to the mean decrease in F1 and the shaded areas are the respective standard deviation. Strikingly, we observe that the constant, blurred, Gaussian and expectation baseline produce competitive results, whereas the uniform and maximum distance baseline tend to generate less discriminative feature attributions.}
\label{fig:average_ablation}
\end{figure*}
%---------- Average Ablation --------------------------------%

\subsection{Data Sets}
A summary of all data sets can be found in Table \ref{tab:datasets}. We standardized (zero mean, unit variance) the continuous features of each data set. Besides, we encoded every non-numeric categorical feature in integers. Finally, we split all data sets, with 80\% of the observations used for training and 20\% for testing.

The Human Activity Recognition (HAR) data set contains sensor signals of a waist-mounted accelerometer and gyroscope from different participants who performed six distinct physical activities. Since KernelSHAP's computation time grows exponentially with the number of features, we performed the ablation test of KernelSHAP on HAR with a stratified sample of 400 observations to enable reproducibility of our results with limited hardware.

The Fraud Detection \cite{dal2014learned} data set contains benign and fraudulent credit card transactions. We used a processed version\footnote{https://www.kaggle.com/mlg-ulb/creditcardfraud} of the data set containing the timestamp and amount of each transaction, along with 28 features generated by a Principal Component Analysis (PCA) of the original transaction details. Since Fraud Detection is a very large data set, we used a random sample of 50,000 observations to compute the experiment in reasonable time. Note that the random sample has approximately maintained the class imbalance of the full data set.

The Communities and Crime \cite{Dua:2019} data set (in the following referred to as Communities) contains socio-economic and law enforcement data on communities in the USA. Our goal was to predict the amount of violent crimes per 100,000 inhabitants. To be precise, we have considered all communities with a crime rate of $>=30\%$ as high risk and others as low risk. We then classified the observations according to these two labels. Note that we removed any feature that had more than 1,000 ($\approx 50\%$) missing values, in order to guarantee valid results.

The Spambase \cite{Dua:2019} data set consists of information about genuine and spam emails.

Finally, COMPAS \cite{angwin2016machine} is an  algorithm that evaluates the risk of recidivism and is known to be biased against black defendants. The COMPAS data set contains the personal information of the defendants. For our experiments, we used a preprocessed version\footnote{https://www.kaggle.com/danofer/compass} of COMPAS.

Note that the blurred, Gaussian and uniform baseline assume continuity of features. Nevertheless, we decided to keep the categorical features of the COMPAS and Communities data set, since we did not perceive a significant change of the ablation tests compared to an evaluation without categorical features.

%--------------------- Comp. Time ------------------------------------%
% \begin{table}[t]
% \caption{\textbf{Computation Time (seconds).} We exhibit the average computation times for generating local attribution scores on a test set (20\% of observations) from every data set. Strikingly, the stochastic baselines are significantly slower, since they involve sampling. The lowest computation time per attribution model is highlighted in bold.}
%     \label{tab:comptimes}
%     \centering
%     \begin{adjustbox}{max width=\columnwidth}
%         \begin{tabular}{lllll}
%         \toprule
%         \textbf{Attribution Model} & \textbf{DeepSHAP} & \textbf{DeepLift} & \textbf{IG} & \textbf{KernelSHAP}\\ 
%         \cmidrule(lr){1-1} \cmidrule(lr){2-5}
%         Constant & \textbf{7.97} & \textbf{0.13} & \textbf{2.87} & \textbf{56.22} \\
%         Maximum Distance & 17.81 & 29.30 & 26.36& 69.36 \\
%         Blurred & 83.25 & 1.04 & 26.14 & 209.90 \\
%         Gaussian & 78.97 & 1.04 & 26.36 & 213.88 \\
%         Uniform & 100.15 & 1.06 & 26.47 & 212.28 \\
%         Expectation & 93.60 & 1.04 & 26.63 & 211.88\\
%         \bottomrule
%         \end{tabular}
%     \end{adjustbox}
% \end{table}
%--------------------- Comp. Time ------------------------------------%

\subsection{Results and Discussion}
As described above, we performed ablation tests to measure the quality of the attributions on different baselines. The computation times of every baseline and attribution model can be found on our Github page.

% Table \ref{tab:comptimes} depicts the average computation time in seconds with respect to each attribution method. Notably, the deterministic baseline methods (constant and maximum distance) have a drastically lower computation time on average, as they do not involve sampling.

Figure \ref{fig:ablation_per_dataset} exhibits the detailed results of every ablation test. Note that if a baseline were to produce discriminative attributions, we would see a sharp decline in F1 scores (since we mask the most important features). In general, we find that all attribution models were very sensitive to the different baselines. In the following, we discuss some of the most important findings. Please note again, that this experimental study is intended as a basis for a more in-depth analysis of baselines on tabular data. Accordingly, the following observations and hypotheses apply in the context of our experiment, but do not necessarily apply in general.

The maximum distance and uniform baselines often struggled to outperform the random baseline. This is in contrast to the results of \citet{sturmfels2020visualizing}, where the uniform baseline achieved competitive results in the ablation test on image data. A possible explanation is as follows: The uniform distribution might not be complex enough to approximate the data generating distribution of heterogeneous tabular data sufficiently well. As a result, the generated baseline values might lie outside the data generating distribution and therefore produce non-discriminative attributions. Likewise, the maximum distance baseline may not be representative of the data. In fact, the maximum distance baseline may return an outlier value. 

On the other hand, the expectation and Gaussian baselines performed well in general. The expectation baseline closely approximates the data generating distribution, since it computes the expectation over a sample of training observations. Since many natural phenomena follow a normal distribution, we expect that the Gaussian baseline also approximates the data generating distribution sufficiently well. These results suggest that feature attributions become more discriminative, the closer a baseline follows the data generating distribution. We leave a detailed analysis for future work.

The constant and blurred baseline showed almost identical performance. Note that by applying a blur filter, we reduce the variance among input features. Accordingly, the blurred baseline approaches a constant value (low variance) the stronger we set the blur effect.

For the extremely imbalanced Fraud Detection data, all baselines performed equally well. Note that Fraud Detection comprises features which were generated in a Principal Component Analysis. Hence, by removing the features that correspond to the first principal components, we remove much of the discriminative information, which would explain the sharp drop in the predictive performance. However, the ablation test does not explicitly consider class imbalances. Hence, these results should be considered with care.

KernelSHAP values are theoretically optimal, but require an exponentially increasing number of feature coalition samples. As \citet{lundberg2018consistent} acknowledge, SHAP values can thus be challenging to compute. Accordingly, the performance of KernelSHAP decreases as the data dimensionality increases, if we can not adjust the number of coalition samples accordingly. This effect can be observed across all baseline methods both in the Communities data set and, to a greater extent, in the HAR data set. Note that the performance of KernelSHAP may be improved by running our experiments on more-advanced hardware.

In summary, our experiments illustrated that the expectation, blurred, constant and Gaussian baselines frequently produce discriminative attributions on tabular data. The uniform and maximum distance baselines generally perform worse in the ablation test, suggesting that they may not be a sensible choice in practice. The average ablation test results in Figure \ref{fig:average_ablation} support these findings. Still, other evaluation methods, such as randomization tests \cite{adebayo2018sanity}, could be considered in the future to substantiate our findings. As indicated in previous work \cite{sturmfels2020visualizing,shrikumar2017learning} and supported by our results, the appropriateness of a baseline method depends strongly on the data distribution at hand. Therefore, it might also be interesting to consider dynamic baseline methods that can be used even if the data generating distribution changes \cite{haug2020learning}. In general, we argue that a conceptual comparison of baseline methods is needed to enable more principled decisions in practice. In this context, the proposed taxonomy of baselines can be an important guideline.

\section{Conclusion}
In this work, we provided a first empirical comparison of common baseline methods for local attributions on tabular data sets. Additionally, we proposed a novel taxonomy of baseline methods. In this way, we complemented existing studies on image data. Our results show that the baseline can have a dramatic impact on the quality of generated feature attributions. In general, we argue that the selection and development of sensible baseline methods should receive more attention in research and practice.

\bibliography{bibliography}

\end{document}